\documentclass[preprint,12pt]{elsarticle}

\usepackage{graphics}
\usepackage{graphicx}
\usepackage{epsfig}

\usepackage{amsmath}
\usepackage{amssymb}
\usepackage{amsthm}

\usepackage{subfigure}
\usepackage{tabularx}
\usepackage{blindtext}
\usepackage{lineno}

\def\etal{\emph{et al}}

\begin{document}
\linenumbers
\begin{frontmatter}
\title{Wavelet-based Scale Saliency}

\author[label1,label2]{Anh Cat Le Ngo}
\author[label3]{Kenneth Li-Minn Ang}
\author[label2]{Guoping Qiu}
\author[label4]{Jasmine Kah Phooi Seng}
\address[label1]{School of Electronic Electrical Engineering, The University of Nottingham, Malaysia Campus}
\address[label2]{School of Computer Science, The University of Nottingham, UK Campus}
\address[label3]{Centre for Communications Engineering Research, Edith Cowan University}
\address[label4]{Department of Computer Science \& Networked System, Sunway University}

\begin{abstract}
Both pixel-based scale saliency (PSS) and basis project methods focus on multiscale analysis of data content and structure. Their theoretical relations and practical combination are previously discussed. However, no models have ever been proposed for calculating scale saliency on basis-projected descriptors since then. This paper extend those ideas into mathematical models and implement them in the wavelet-based scale saliency (WSS). While PSS uses pixel-value descriptors, WSS treats wavelet sub-bands as basis descriptors. The paper discusses different wavelet descriptors: discrete wavelet transform (DWT), wavelet packet transform (DWPT), quaternion wavelet transform (QWT) and best basis quaternion wavelet packet transform (QWPTBB). WSS saliency maps of different descriptors are generated and compared against other saliency methods by both quantitative and quanlitative methods. Quantitative results, ROC curves, AUC values and NSS values are collected from simulations on Bruce and Kootstra image databases with human eye-tracking data as ground-truth. Furthermore, qualitative visual results of saliency maps are analyzed and compared against each other as well as eye-tracking data inclusive in the databases.
\end{abstract}


\end{frontmatter}

\section{Introduction}
\label{sec:intro}

A few centuries ago, Neisser proposed a fundamental theory about the human visual attention system including pre-attentive and attentive stages \cite{Neisser1967} in his psychology studies. However, his work was unknown to machine vision scientists until David Marr \cite{Marr1982}, a neurologist, proposed a neurology-based computational model for Neisser's theory. The computational model includes a feature extraction stage followed by perceptual grouping stage. Though the model was practically limited and rarely implemented, it inspires and provides framework for several later computational models. Among them, Itti model \cite{itti1998model} holds significant influence and provides a standard in the research field. Itti models feature extraction as center-surrounds operator, a property of visual cortex; while, perceptual grouping and attentive region assessment are due to Proto-Object generation \cite{Walther2006} and Winner-Take-All network \cite{itti1998model}.
After center-surrounds operations in multi-scale levels were proposed for construction of conspicuity and saliency maps by Itti \etal, other theories like Graph-based Visual Saliency \cite{Harel2007}, and Spectral Residual Saliency \cite{Hou2007} were brought in to produce more meaningful saliency maps \cite{Harel2007} as well as reduce the computational complexity \cite{Hou2007}. These saliency models assume that human vision systems may behave like random-walk processes \cite{Harel2007} or follow statistical property of natural images \cite{Harel2007}. 

Without making such strong assumption, Kadir \cite{Kadir2001} and Gilles \cite{Gilles1998} initiated information-based saliency map with their work on pixel-based scale saliency (PSS). Other information-related saliency research rapidly gained pace with Niel Bruce's An Information Maximization (AIM) theory \cite{Bruce2006} and Danash Gao's Discriminative Information Saliency (DIS) \cite{Gao2007}. Furthermore, the information-based spatial-temporal framework (ENT) \cite{AnhCat2012} \cite{Qiu2007} extends and fastens the models from still images to the dynamic video context. 

Information-based saliency approaches all motivated from the assumption that human attention could be attracted to spatial location accompanied with highly informative content. From signal coding, compression and self-information theory, an event has more information when it appears to be structural and rare. Though based on similar concepts, each method has its own information estimation approach on different type of descriptors. Generally, those approaches can be characterized according to their choices of descriptors and calculation methods. For examples, PSS \cite{Kadir2001} and ENT \cite{Qiu2007} utilize pixel-value descriptors; meanwhile, AIM \cite{Bruce2006} and DIS \cite{Gao2007} emphasizes on the alternative basis-projection descriptors, ICA bases and Wavelet bases consecutively. In accordance with information measurement, ENT and PSS employ the popular Shannon entropy estimated by histogram construction or Parzen kernel. AIM estimates self-information through neural-network training on patches of natural images. Decision-theory based DIS has its discriminative information from classifying descriptors into center or surrounds classes. Noteworthy, PSS is so far the only approach accumulating information of both descriptors and their structure. However, PSS employs pixel-value descriptors and isotropic circular sampling, which might hinder its performance in term of accuracy due to failure in extracting popular oriented features in natural images as well as speed due to the curse of dimensionality in information estimation.

The limitation of PSS sparked discussion for alternative solutions by Kadir \etal \cite{Kadir2001} \cite{Kadir2003}. Deploying basis-projected descriptors in place of pixel-based descriptors not only boosts practical performance of scale saliency but as well provides deeper theoretical understanding of scale saliency and data multi-scale structural information. Moreover, the extension would make scale saliency the first-ever method capable of using both pixel-value descriptors (PSS) and basis-projection descriptors. Wavelet elements are preferred as alternative basis in this paper; therefore, the proposed method is named Wavelet Scale Saliency (WSS). In order to clearly explain extension from pixel-based descriptors to basis-projection descriptor, we organize this paper in following sections. Section \ref{sec:scale_saliency} gives overview about scale saliency and its main idea. The next section \ref{sec:tfsdomain} explains the rationale behind usage of time-frequency domain instead of time-domain only for visual saliency; meanwhile, sections \ref{sec:distribution1d}, \ref{sec:distribution2d} elaborates statistical distribution and correlation of time-frequency descriptors. As wavelet is chosen as time-frequency basis, section \ref{sec:waveletTransform} gives background information about four types of wavelet transforms considered in this study: discrete wavelet transform (DWT), discrete best basis wavelet packet transform (DWPTBB) as well as two quaternion wavelet transforms QWT and QWPTBB. Accordingly, there are four time-frequency descriptors representing time-frequency domain slightly different from each other. Moreover, each descriptor depends on a particular morphological shape of its own mother wavelet. All details about properties of those descriptors are organized in section \ref{subsec:descriptors}. Along with new descriptors, suitable mathematical models of feature-space and inter-scale saliency estimation are derived in section \ref{sec:calculation}. Moreover, the mathematical derivation unveils strong relation between WSS with another state-of-the-art Bayesian Surprise Saliency (BSS) \cite{Baldi2010}. Beside theoretical evaluation, simulations on Neil Bruce image database \cite{Bruce2009a} and Kootstra image database \cite{Kootstra2011} are carried out in order to compare quantitatively the proposed WSSs with different basis-projection descriptors, the original PSS and the ITT model. Furthermore, qualitative analysis on particular images provides better details about responses of the proposed methods with different types of scenes. It is possible that performance of saliency methods depends on image content. All results and discussion are detailed in the section \ref{sec:results}. Finally, the conclusion \ref{sec:conclusion} summarizes our main contributions and future research directions.

\section{Scale Saliency}
\label{sec:scale_saliency}

To get a hold of what exactly is a scale saliency; a few fundamental principles of original scales saliency are reviewed. Scale saliency utilizes maximum feature-space entropy weighted by its inter-scale dependency across scales as saliency values; furthermore, it argues that information measurement might be data-driven pivot for human visual attention. Its mathematical model is summarized as follows.
\begin{eqnarray}
	Y_{D} \left( s_p,\vec{x} \right) &{} = {}& H_{D} \left( s_p,\vec{x} \right) W_{D} \left( s_p,\vec{x} \right) \label{eqnarr1:sl} \\
	H_{D} \left( s_p,\vec{x} \right) &{} = {}& -\int_{d \in D} p \left( d,s_p,\vec{x} \right) \log_2 \left( p(d,s_p,\vec{x}) \right) dd  \label{eqnarr1:hd} \\
	W_{D} \left( s_p,\vec{x} \right) &{} = {}& s\int_{d \in D} \vert \frac{\delta p \left( d,s_p,\vec{x} \right)}{\delta{s}} \vert dd \label{eqnarr1:wd} \\
	s_p &{} = {}& \lbrace s \vert \frac{\delta H_D \left(s,\vec{x}\right)}{\delta s} = 0;\frac{\delta^2 H_D \left( s,\vec{x} \right)}{\delta s^2} < 0 \rbrace \label{eqnarr1:sp}
\end{eqnarray}
Feature-space saliency, ($H_D$) in the equation \ref{eqnarr1:hd}, is measured by its Shannon entropy of pixel-values descriptor $(d)$ at a specific scale or sampling window size $(s)$ for each image location $(\vec{x})$. Shannon entropy is chosen since it satisfies fours over five criteria of multi-scale entropy filtering \cite{Starck1999}. The last criterion actually requires structural correlation from information estimation, which is apparently not considered in Shannon entropy. However, inter-scale saliency, $(W_D)$, actually fulfils this requirement, and it is estimated at every location by total variation of descriptors' probability distribution function ($PDF$) across scales. Then, the scale $(s_p)$ at which most significant information should be found; it is actually the maximum point of the scale-entropy concave curve in the equation \ref{eqnarr1:sp}. Finally, the overall saliency is stated mathematically as the equation \ref{eqnarr1:sl} in accordance with the definition of scale saliency. Lets apply the concept of scale saliency on a general form of signal $R(x_0,s_i) =\lbrace I(\vec{x_0},s_i) + N(\vec{x_0},s_i) \vert i = 1 ... n \rbrace$, where $I_{\vec{x_0},s_i}$ is ideal noise free signal, $R_{\vec{x_0},s_i}$ is the measured signal with noise $N_{\vec{x_0},s_i}$ at specific location and scale $(\vec{x_0},s_i)$. Assumed no dependencies between noise and the ideal signal, the estimated entropy is $H_{D}\left(R_{\vec{x_0},s_i}\right) = H_{D}\left(I_{\vec{x_0},s_i}\right) + H_{D}\left(N_{\vec{x_0},s_i}\right)$. Assumed that noise $PDF$ are scale-invariant, the equation \ref{eqnarr1:wd} implies that inter-scale saliency measure is purely dependent on variation of useful signal $\Delta_{s_i} H_{D}\left(I_{\vec{x_0},s_i}\right)$ and not affected by variation of noise $\Delta_{s_i} H_{D}\left(N_{\vec{x_0},s_i}\right) = 0$. This briefly explains basic motivation behind scale saliency work; further mathematical analysis and experiments results can be found in \cite{Kadir2001} \cite{Kadir2003}.

The original scale saliency \cite{Kadir2001} uses pixel-value descriptors which are simple, intuitive, and straight forward interpretation of image data. Moreover, its combination with circular sampling window provides isotropic information analysis, independent of any morphological shape inside sampled regions. Nevertheless, its drawbacks are be susceptible to noise, require high computational cost and cause significant bias in entropy estimation. So far, histogram construction and approximated Parzel kernel are two popular parameter methods for constructing pixel-value descriptors' $PDF$ and estimating entropy. Entropy bias and speed performance in those mentioned methods greatly depend on manual tuning of histogram numbers of bins or Parzel size kernel; in addition, they as well restrict extension of scale saliency to higher dimensional data. Suau \cite{Suau2009} overcomes these problems by bypassing $pdf$ construction stage and estimating PSS by multivariate-data-adaptive information estimation technique \cite{Stowell2009b}. In spite of its fast computation for multivariate data, the non-pdf approach hinders the inter-scale saliency process which directly depends on $PDF$s \ref{eqnarr1:wd}. It is solved by adapting set-theory based elegant solutions of Kadir \cite{Kadir2003} for inter-scale saliency $W_D$ computation into kd-tree structure. However, the solution is not intuitively and mathematically related to the information-based frame-work. That motivates us develop ($WSS$), a more coherent information-based scale saliency with sub-band energy descriptors, as solutions for all these short-comings of PSS.
\section{Time-Scale-Frequency}
\label{sec:tfsdomain}
A well-known computational model of visual attention is first mentioned in Koch and Ullman's publication \cite{Koch1985}. After that, several other models are proposed; however, they are usually over-complex and not biologically plausible. The disadvantages might be due to pixel representation utilized in many early visual attention algorithm. To overcome these problems systematically, Urban \cite{Urban2010} has investigated strong constraints to keep computational complexity within an acceptable range for possible real-time implementation. These constrains are drawn from evidences of psychological experiments which shows that images could be analyzed in psycho-visual channels at least in TV-viewing condition \cite{Watson1987}. In other words, visual data could be further analysed into channels and sub-bands instead of being used in raw pixel format. Furthermore, the channels can be effectively characterized by separated frequency bands and orientation ranges of wavelet analysis \cite{Senane1995}. Lets assume visual active areas of brains can deploy some 9/7 Cohen-Daubechies-Feauveau (CDF) wavelet transform operators; then, it results in multi-scale pyramid composed of oriented contrast maps with limited frequency range and low-resolution image. For each level of wavelet decomposition, there are four channels: (i) sub-band 0 is approximated image after filtered with many low-pass blurring kernels; (ii) sub-band 1 extracts horizontal frequencies corresponding to vertical edges of images (iii) sub-band 2 contains frequencies and features along two diagonals of image frames. (iv) sub-band 3 prefers vertical frequencies mapping to horizontal features form images. Natural scenes are full of horizontal, vertical or two diagonals features; therefore, human visual perception seems to prefer those dominant features. Besides oriental constraints, visual acuity is another visually perceptual limit. Normally, human fovea could decompose and process details above its limit visual acuity (1.5-2 degrees of visual angle). It lasts in frequency range: 0.7-0.5 pixels per degree,or 0.33-0.25 cycles per degree. This range is nearly resembled by the last level low-resolution version of images in usual wavelet decomposition. Each decomposed level are generated by moving kernels with different window size to any image positions. Spatial frequency of other wavelet analysis levels, varying  in accordance with analyzing depths, is shown the following table \ref{tab:wavLevFreRng}

\begin{table}
\begin{tabular}{ccccccc}
\hline 
Depth & 0 & 1 & 2 & 3 & 4 & 5\tabularnewline \hline
Frequency range \\ (cycles per degree) & 10.7-5.3 & 5.3-2.7 & 2.7-1.3 & 1.3-0.7 & 0.7-0.3 & 0.3-0.2\tabularnewline
\hline 
\end{tabular}
\caption{Wavelet Levels vs Frequency Range}
\label{tab:wavLevFreRng}
\end{table}

Spectral energy are usually employed as spectral signature for image collections or individual images. Urban \etal \cite{Urban2010} analyses different sets of images belonging to four different semantic categories: coast, mountain, street and open-country. Interestingly, Fourier spectrum of each category possess distinguished shape and frequency range, significantly different from each other \cite{Urban2010}. In other words, each general spectral profile and associated distribution histogram of image classes have unique energy distribution. This distribution is proportional to distance $d$ from mean magnitude spectrum normalized by the standard deviation of the category. 
\[
	d(s)=\frac{1}{nm}\sum_{n,m}|s(n,m)-AS(n,m)|
\]
where $AS(n,m)$ is average spectrum. Carefully observing the spectral profiles could give distinguishing clues for each semantic scene. For example, \textit{"Coastal"} scenes are dominated with horizontal features; therefore, its spectral profiles stretch along vertical axes. Furthermore, spectral profile of \textit{"OpenCountry"} categories is biased toward two upper and lower spectrum.  Though almost similar to the \textit{"OpenCountry"} profile,  spectrum of \textit{"Street"} images includes more types of features from artificial environment beside horizon-oriented details. Therefore, the diamond of image spectral profile of  \textit{"Street"}  becomes more significant horizontally. \textit{"Mountain"} categories with its random scenic details have isotropic spectrum while scenes of streets filled with artificial objects have spectrum stretched in both horizontal and vertical axis. From Urban's research, spectral energy distribution seems to be important clues for visually perceptual system of human beings.

Beside image classification, the spectral distribution signature is as well useful in visual attention and early visual process. Such energy distribution becomes differentiable clues for features across scales. Spectral profiles of image feature at a particular scale would help differentiate itself from directly upper and lower scale. Lets do an imaginary experiments with a single square input signal $x(t)$ defined as follows.
\[
x(t)=\begin{cases}
1 & t_{1}\leq t\leq t_{2}\\
0 & t<t_{1}\vee t>t_{2}
\end{cases}
\]

If $x(t)$ is filtered by a kernel $F()$ with kernel size ( 1-D kernel width ) $W=\Delta T$ and $W$ is much smaller than non-zero period of the given square signal $\Delta T\lll\vert t2-t1\vert$, the response
will be just two impulse function at $t1,t2$.
\[
F(x(t))=\begin{cases}
1 & t=t_{1}\wedge t=t_{2}\\
0 & t\neq t_{1}\vee t\neq t_{2}
\end{cases}
\]
For 2-D signal or image context, the above operation corresponds to a classic edge detection phenomena. Though edges and structures plays important roles in visual perception, their information does
not sufficiently represent the whole natural scenes. Natural images are rich of other features like texture, flat regions, etc beside edges and corners.  As mentioned before, image features can be interpreted  in terms of energy distribution. For example, edges are the places of high energy concentration, homogeneous flat regions do not contain much energy while textures, hybrid of edges and flat regions, contains certain amount of energy . If only one window size is used in the analysis, significant responses only come from features or objects which happen to fit into that window size. The other useful features with inappropriate size in accordance with the filter could not be extracted. Therefore, a multi-scale's approach is extremely necessary in order to identify suitable sizes of kernels or fuse features from different scales together. When mother wavelet is chosen as filtering kernels, window size becomes equivalent to frequency range in the table \ref{tab:wavLevFreRng}, and choosing adaptive frequency ranges is important computation task for spatial feature extraction. Inspired by such fundamental query in computer vision, this paper tries to contribute a little insight about how spectral density distribution can characterize features at each scale and how the frequency range of processing can be appropriately chosen for multi-scale feature representation. From the multi-scale features and appropriate scale selection, we can develop computation saliency methods capable of highlighting salient features across scales by using spectral energy distribution.

\section{Time-Scale-Energy}
\label{sec:distribution1d}
PSS estimates information from pixel values, time-domain descriptors by constructing normalized histogram of pixel values as probability distribution.
\[
	p_{h}(d) = \frac{n_d}{N}
\]
where $p_{h}$ is probability of descriptor, the ratio between number of pixels with $d$ descriptors and total image pixels N. Lets use square of pixel $d^2$ as weights
\[
	p_{e}(d) = \frac{n_d*d^2}{\sum_{D}(n_d*d^2)} = \frac{\rho_x}{E_x} = \frac{\int\limits_{-\infty}^{\infty} \left( x(t)\delta(x(t)-d) \right)^2 dt}{\int\limits_{-\infty}^{\infty} x(t)^2 dt}
\]
Normalized weighted-histogram $p_{e}(d)$, of signal $x(t) \in L^2(R)$ can obviously be interpreted as $\rho_x$ energy density of descriptors $d$ in time domain. By the isometric property of the Fourier transform, the $PDF$ of energy density distribution can be expressed in frequency domain as well.
\[
	p_{e}(\hat{d}) = \frac{\rho_{\hat{x}} }{E_{\hat{x}}} = \frac{\int\limits_{-\infty}^{\infty} \hat{x}(f)^2\delta(\hat{x}(f)-\hat{d}) df}{\int\limits_{-\infty}^{\infty} \hat{x}(f)^2 df}
\]
Or in joint time and frequency domain.
\[
	p_{e} =  \frac{\rho_ {\bar{x}}}{E_{\rho_{\bar{x}}}} = \frac{\rho_{\bar{x}}(t,f)}{\int\limits_{-\infty}^{\infty} \int\limits_{-\infty}^{\infty} \rho_{\bar{x}}(t,f) dt df}
\]
where
\[
	\rho_{\bar{x}} = \left| \int\limits_{-\infty}^{\infty} x(\tau)g^{*}_{t,f}(\tau)d\tau \right|^2
\]
where $\rho_{\bar{x}}$ is energy density in joint time-frequency representation. Pure time descriptors have perfect localization in time, no localization in frequency , and vice versa for frequency descriptors. Both extreme time or frequency descriptors make interpretation of constructed $PDF$, and estimated information difficult to explain. Therefore, it is necessary to find a representation of $g^{*}_{t,f}(\tau)$ which describes spectral density of local energy. For example, Short-Time Fourier Transform (STFT) is the first-known transform capable of generating spectrogram, a graphical representation of local signal energy in time-frequency plan.
\[
	\rho_x(t,f) = \left|{ \int\limits_{-\infty}^{\infty} x(\tau) h^*(\tau - t) e^{-2j\pi f\tau} d\tau}\right|^2
\]
STFT identifies spectral density as well as local energy density or information in a short-time period of the signals. However, it does not much benefit scale saliency unless scale parameters are actually considered as in signal description on phase-space. Fortunately, in recent years, alternative scale-based representation, called wavelet-transform (WT), has been widely addressed among signal processing community, and its fundamental idea is replacing the frequency shifting operation $e^{-2j\pi f\tau}$ by a time (or frequency) scaling operation $\psi(\frac{t-\tau}{a})$, a basic wavelet kernel. Consequently, the energy density in WT framework is formulated as follows.
\[
	\rho_x(t,a) = \left|{ \int\limits_{-\infty}^{\infty} x(\tau) \psi(\frac{\tau - t}{a}) d\tau}\right|^2
\]
WT coefficients, $\rho_x(t,f)$, averagely measure spectral density of frequency sub-bands, a short range of frequency, in a short period of time. Characteristics of the time-frequency window are specified by two main parameters, time-shift $\tau$ and scale $a$. As derived from short-time spectral representation, a spectrogram, by utilizing scale operations, its energy density distribution is called scalogram. Given time-scale space, the total signal energy can be rewritten as follows.
\[
E_{\rho_x} = \int\limits_{-\infty}^{\infty} \int\limits_{-\infty}^{\infty} \rho_{x}(t,a) dt da
\]
and probability of time-scale descriptors can be specified with scale parameters
\begin{equation}
	P_{e}(t,a) = \frac{\rho(t,a)}{E_{\rho_x}}
	\label{equ:pdf1d}
\end{equation}
Generally, the wavelet transform can help generate frequency sub-band coefficients, square of which over total energy are density distribution of that sub-band in time-scale space.

\section{Wavelet Transform}
\label{sec:waveletTransform}
After discussing about usefulness of time-frequency-scale representation in the section \ref{sec:tfsdomain} and its corresponding time-frequency energy distribution, we recognize that wavelet-representation would be ideal candidates for our investigation into energy density distribution and other statistical property across multiple scales of natural images. During the quite short history of wavelet analysis, this research fields have been very fruitful and there are several analysing techniques with wide range of characteristics. In this paper,  only standard techniques such as discrete wavelet transform (DWT), discrete wavelet packet transform (DWPT), quaternion wavelet transform with best-basis(QWTBB), and quaternion wavelet packet transform with best-basis. are deployed as  possible descriptors. In this subsection, we first look into discrete and real wavelet and wavelet packet transform with best basis (DWT,DWPTPP) for its theoretical background; then QWT and QWPTBB, quaternion versions of two prior wavelet transforms, are considered.

\subsection{Discrete Wavelet Transform}
\label{subsec:dwt}
Though wavelets were firstly introduced in the early 20th century by Alfred Harr, they are only developed rapidly much later. Only until recently, they have been widely employed in many computer vision problems such as image or video de-noising, enhancement, coding, and pattern classification \cite{Unser2011a,Do2005,Chan2008} . Signal analysis for frequency components can be achieved by Fourier transform (FT) but FT does not provide suitable tool for time-frequency analysis of images. Short-time Fourier Transform (STFT) is an extension from FT approach for analysing local frequency analysis at a short period of time \cite{Mallat1999}. Noteworthy that, STFT can be used for taking the spatial interval in 2-D signal instead of time period in 1-D type since there is no time dimension for still images. However, STFT utilizes fixed window kernels for every data blocks across input signals; this property make STFT less suitable for complex signal analysis, especially signals with strong semantic structures appearing across multiple scales. In other words, STFT only succeeds with signals whose features are embedded in fixed definite temporal or spatial regions or there is prior knowledge about a suitable size of window kernel for STFT processing. Without the above conditions, STFT would totally miss signal features. Theoretically, the disadvantage can be avoided by employing STFT with multiple kernel sizes; however, it raises up another issues such as what range of sizes would be chosen to optimally extract useful features with reasonably computational effort.

Problems of STFT in analyzing local frequency have motivated development of multi-scale wavelet techniques for better local frequency representation. Since limitations of STFT is due to fixed-size processing windows, wavelet analysis deploys multi-resolution filter-banks on input signals. As illustrated in figure \ref{fig4:dwptIllus}, 1-D signals are decomposed into low-pass and high-pass components. In case of 2-D input signals, the filtered outputs are four sub-bands: low-low, high-low, low-high, and high-high in regards of processing orientation. Intuitively, 2-D signals analysis includes
row-wise 1-D analysis followed by column-wise 1-D analysis or vice verse. With respect to processing direction, high-low sub-band tends to extract horizontal features, low-high sub-band prefers vertical
features, high-high sub-band detects diagonal features, and low-low are approximated version of original signal by inverse dyadic scale. Lets assume that input signals are two-dimensional grey-scale image $f(x,y)$, and the scaled mother wavelets have following mathematical form$\psi_{s,i}(x,y)=2^{s}\psi(2^{s}x,2^{s}y)|_{i=\{v,h,d\}}$ for
vertical, horizontal and diagonal sub-bands and a scaling function $\phi_{S}(x,y)=2^{S}\phi(2^{S}x,2^{S}y)$ for low-resolution signals
with $s\geq S$. Then, we can represent any images $f(x,y)\in L_{2}(\mathbb{R})$ 
as. 
\[
f(x,y)=\sum_{x,y}c_{S}(x,y)\phi_{S}(x,y)+\sum_{i=\{v,h,d\}}\sum_{x,y,s\geq S}d_{s,i}(x,y)\psi_{s,i}(x,y)
\]
where \[c_{S}(x,y)=\int f(x,y)\phi_{S}(x,y)dxdy\] 
	  \[d_{s,i}(x,y)|_{i=\{v,h,d\}}=\int f(x,y)\psi_{s,i}(x,y)dxdy\] 
	  
$c_{S}(x,y)$ and $d_{s,i}(x,y)$ are scaling coefficients and wavelet coefficients from vertical, horizontal and diagonal sub-bands. The parameter S represents the lowest analyzing depth while s is higher decomposing levels in multiple scale-space framework. As mentioned before, 2-D DWT can be obtained by tensors products of 1-D DWT when the orginal image $f(x,y)$ is analyzed along two dimensions $x$ and $y$ separately. As a result, the scale function $\phi(x,y)$ is approximated as $\phi(x)\phi(y)$ and filter-banks of three directional
sub-bands are $\phi(x)\psi(y),\psi(x)\phi(y),\psi(x),\psi(y)$. In the figure \ref{fig4:dwptIllus}, discrete wavelet transforms are carried on the sample image in the left hand-side. On the right-hand side contains decomposed results by two levels with three distinctive sub-bands and a down-sampled version of the original image.
\begin{figure}[!htbp]
	\centering
	\subfigure[Sample Image]{\includegraphics[width=0.45\textwidth,natwidth=512,natheight=512]{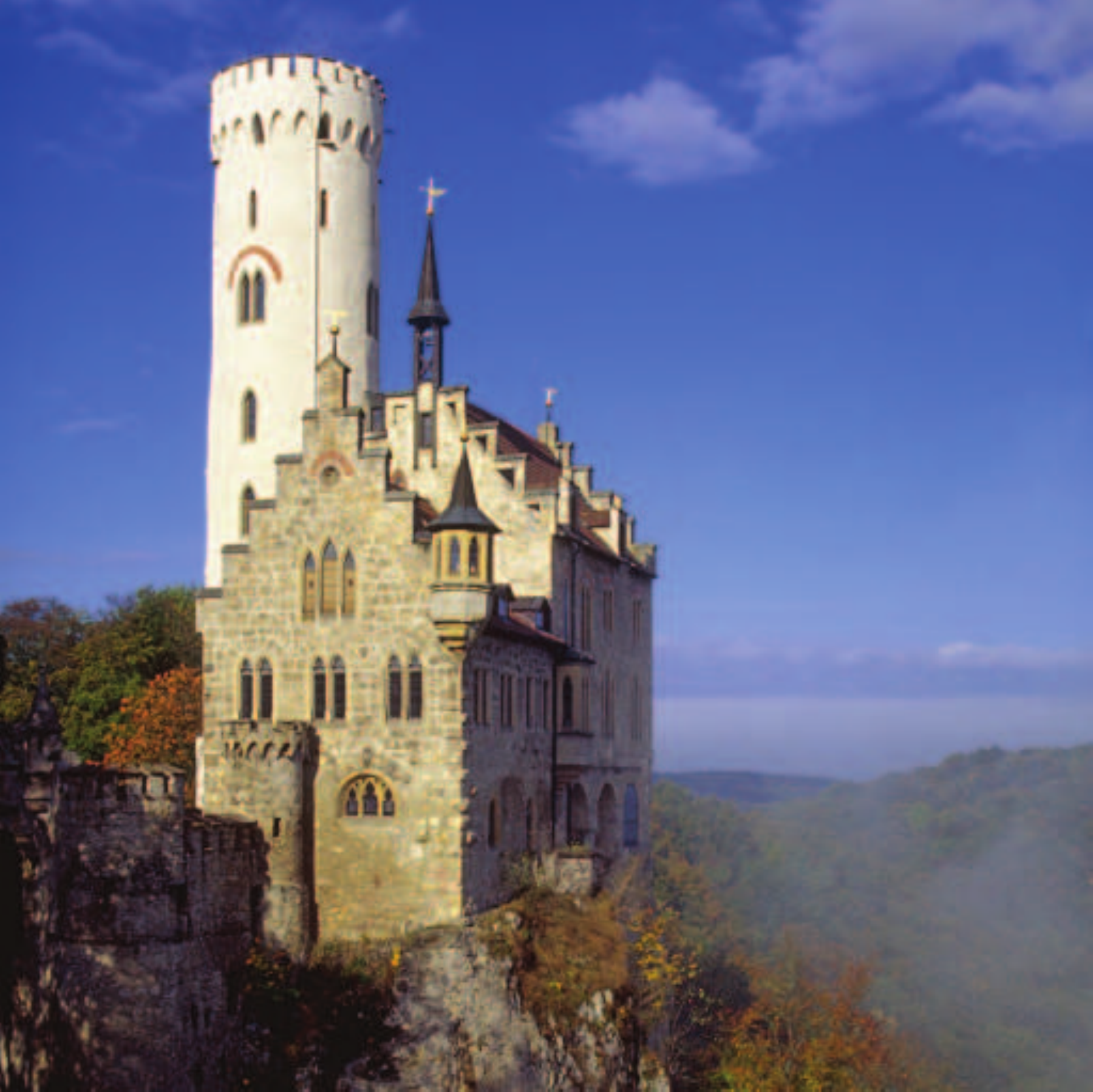}\label{fig3:sample}} \qquad
	\subfigure[Wavelet Decomposition]{\includegraphics[width=0.45\textwidth,natwidth=512,natheight=512]{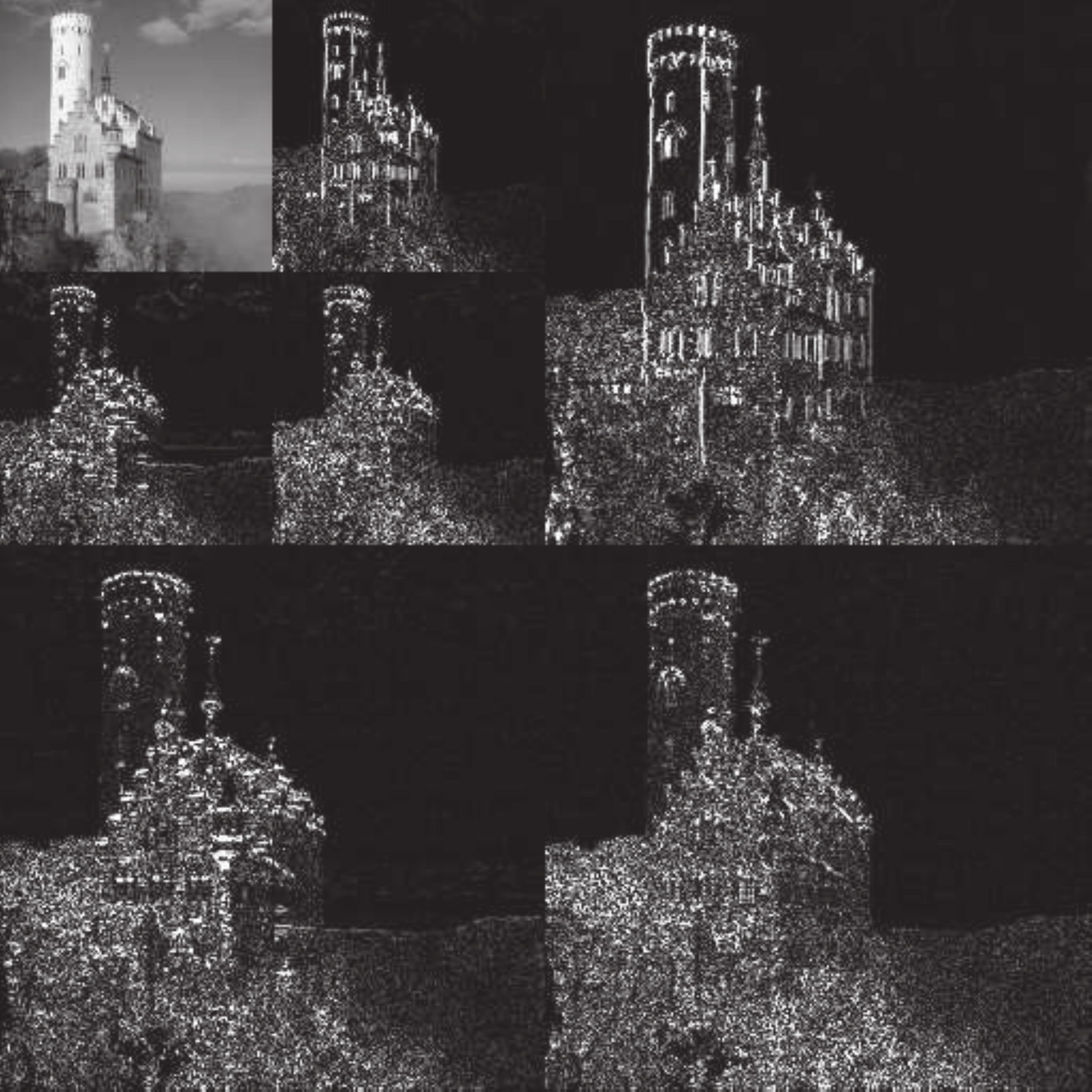}\label{fig3:coefficients}}
\end{figure}

Noteworthy that, the real-wavelet transform like DWT suffers from shift-variance, a small shift in the signal can greatly change magnitudes of wavelet coefficients around singularities. Furthermore, it has
no phase to embed signal location information therefore aliasing effects would be introduced into recovery process. These issues need seriously considering whenever the discrete real wavelet transform is employed. Therefore, modelling statistical property of DWT coefficients' magnitude across scales might request extra investigation with those draw-backs in mind. Further arguments and details about this matter of DWT descriptors will be discussed in the section \ref{sec:scale_saliency}.
\begin{figure}[!htbp]
	\centering
	\includegraphics[width=0.75\textwidth,natwidth=329,natheight=285]{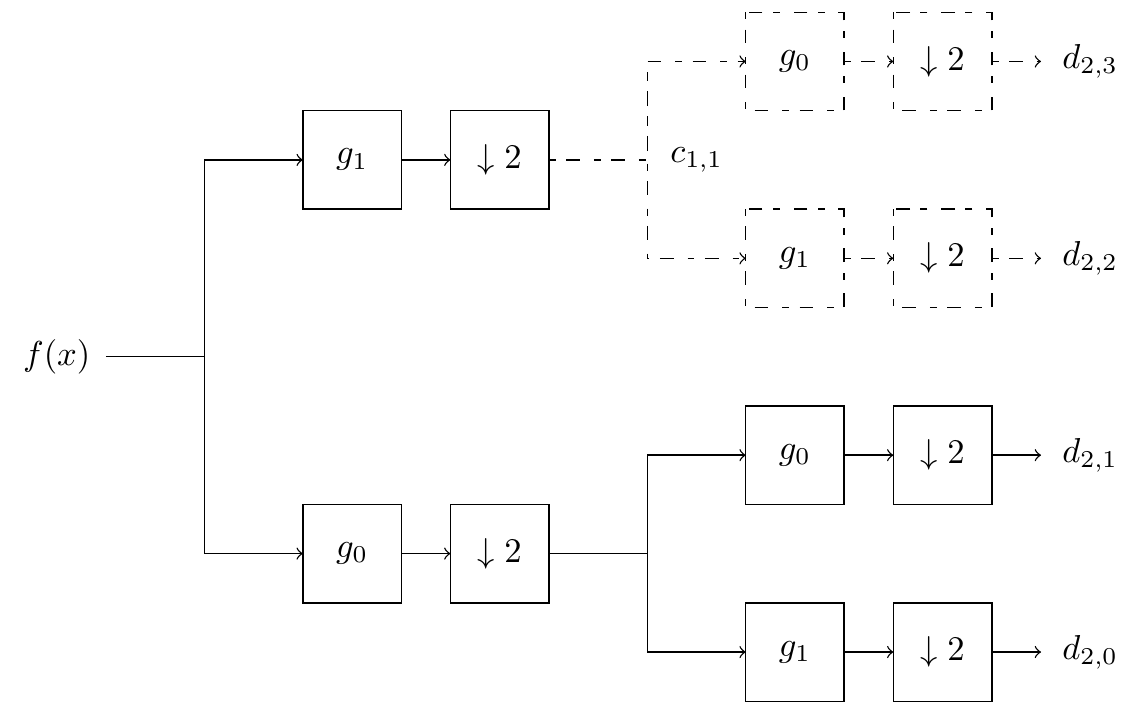}
	\caption{Wavelet (solid line) \& Wavelet Packet Decomposition (solid and dash lines) for 1-D signal}
	\label{fig4:dwptIllus}
\end{figure}
\subsection{Discrete Wavelet Packet Transform}
\label{subsubsec:dwptbb}
Well-known DWT can be computed efficiently by an orthonormal FIR conjugate quadrature filter banks $g_0,g_1$ including analysis low-pass and high-pass filters ( denoted $\phi_{S}$ and $\psi_{s,i}$ respectively. The low-pass coefficients $c_{S}(x,y)$ are decomposed recursively for a number of levels, and inverse discrete wavelet transform is calculated by an inverse filter bank. The extension from normal wavelet transform (DWT) to wavelet packets transform (DWPT) is straightforward by an additional step at each processing level. Instead of decomposing only low-pass coefficients in  the Low-Low sub-band for 2D input signals, the transform performs decomposition of high-pass coefficients in Low-High, High-Low and High-High sub-bands as well. As a result, all coefficients of DWPT  can be neatly arranged in a binary tree and addressed as follows.
\[
d_{s,i}(x,y)\,\,,s\in[0,S],i\in[0,4^{s}-1]
\]
where $s$ is a analysis depth in the tree, $S$ notes the deepest decomposed level, and i is the node index in this depth. With regards to other representations, wavelet packets have advantages
in their adaptability to varying statistical structure. Unlike Fourier Transform with one fixed-size base or normal wavelet transform with a fixed number of bases, we may search the ``best''
orthonormal bases from dictionary of basis acquired after wavelet package decomposition. This idea is initially proposed by Coifman \etal \cite{Coifman1992} mainly for signal compression. Therefore, this ``best'' basis is the best in terms of compressing ratio which often desires sparsest representation. In other words, input signals can be characterized by few large coefficients. Supposed the whole best basis operation is denoted as $B^{2}$ which exhaustively goes through the whole binary tree to look for locations of a set basis with parameters $\mathbf{(s,i)}$ such that the there is a minimum amount of uncertainty measured by Shannon entropy. More details can be found in Coifman's works \cite{Coifman1992}, and $B^{2}$ can be summarized mathematically as follows.
\[
B^2(d_{s,i}): (\mathbf{s},\mathbf{i})=argmin_{s,i}(\sum H(d_{s,i}))
\]
Noteworthy that, sometime ``brute-force-attack" every branch of the tree is not possible or feasible due to intensive requirement of computational power. Fortunately, there exist fast algorithms to implement the best basis for given signals. Then, the optimum time-frequency representation can be achieved by tilting the time-frequency plan in accordance with best-basis algorithms. Though the representation may be optimally sparsest in time-frequency domain, whether sparseness of features suitably matches performance of human visual attention is still a question to be answered. To rectify the matter, experiments have been carried out and performance comparison between the DWT case and DWPTBB case is reported in the section \ref{sec:results}.

\subsection{Quaternion Wavelet Transform}
\label{subsubsec:qwt}
Like wavelet packet transform in the previous discussion, Quaternion Wavelet Transform (QWT) is extended and enhanced to eliminate shift-variance problems from real discrete wavelet transform (DWT). Though there are a few different definitions and implementations of Quaternion Wavelet \cite{Bulow2001}, the QWT implementation in this paper is inspired by Chan's research \etal \cite{Chan2008}.  Some backgrounds about complex wavelet transform (CWT) with its implementation dual-tree complex wavelet transform (DT-CWT) Kingsbury \etal \cite{Kingsbury1998} need reviewing before the QWT can be explained and discussed. In discrete wavelet transform, 2-D DWT can be considered as concatenation of two consecutive 1-D DWTs. Though the same process does not exactly happen in 2-D DT-CWT or QWT straightforwardly. A similar concept is used for easily explaining how QWT can be achieved. It means 1-D DT-CWT will be elaborated first; then, we will discuss about 2-D QWT signals and how it may handle processing along different orientations and sub-bands.

\begin{figure}[!htbp]
\centering
\includegraphics[width=0.75\textwidth,natwidth=258,natheight=234]{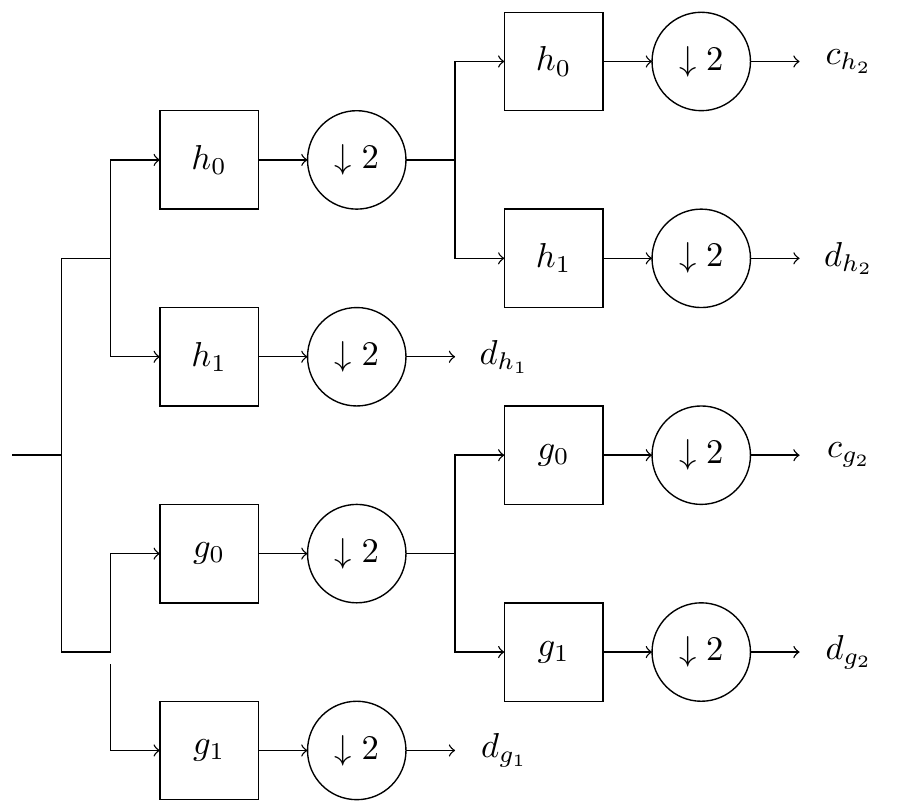}
\caption{Illustration of DTCWT}
\label{fig:dtcwt_illus}
\end{figure}

Real DWT have well-known drawbacks in terms of shift-invariance and phases to encode coefficient locations. Kingsbury \etal \cite{Kingsbury2001} reckons problems and proposes an dual-tree
CWT as a specific solution. Rationale of the dual-tree approach is usage of complex numbers for wavelet coefficients which directly tackles one of two DWT's dragging problems. Complex extension of wavelet transform makes phase extraction from  wavelet coefficients possible since complex wavelet transforms have both real and imaginary values unlike real DWT with only one real value for each coefficient. Real and imaginary components of the dual-tree CWT are generated by two sets of wavelet and scaling functions $\psi_{h},\psi_{g}$and $\phi_{h},\phi_{g}$. Moreover, filter-banks $h_{0},h_{1}$and $g_{0},g_{1}$ have to be independent and orthogonal as shown in the figure \ref{fig:dtcwt_illus}. 

The notations $\phi_{h}(x)$ and $\psi_{h}(x)$ are denoted for scaling and wavelet functions corresponding to filter-banks $h_{0},h_{1}$.  In addition,$c_{h_{S}}$and $d_{h_{s}}$with $s\leq S$ denotes first set of DTCWT coefficients. Similar notations are used for the second set of scaling and wavelet functions $\phi_{g}(x)$ and $\psi_{g}(x)$ with filter banks $g_{0,}g_{1}$ and according coefficients $c_{g_{S}}$and $d_{g_{s}}$ with $s\leq S$. Wavelet functions $\psi_{h}(x)$ and $\psi_{g}(x)$ forms two binary trees in figure \ref{fig:dtcwt_illus} and at leaves of each tree is the real and imaginary parts of a complex analytic wavelets.
\[
\psi^{(c)}(x)=\psi_{h}(x)+j\psi_{g}(x)
\]
Moreover, the imaginary wavelet $\psi_{g}(x)$ is 1-D Hilbert Transform of the real-wavelet $\psi_{h}(x)$: 
\[
\psi_{_{g}}(x)={\cal HT}(\psi_{h}(x))
\]
Any complex wavelet coefficient are formed by wavelet coefficients of two other real wavelet transforms; therefore, this combination generates a 2 x redundant tight frame. This redundancy in complex wavelet frames prevents non-oscillating magnitudes of coefficients around singularity points as well makes the transform near-shift invariant. Furthermore, there is no energy ( or little energy in practice ) in the negative region of frequency because of a relationship between two wavelet functions $\psi_{h}$ and $\psi_{g}$.
\[
\Psi_{g}(\omega)=\begin{cases}
-j\Psi_{h}(\omega) & \omega>0\\
j\Psi_{h}(\omega) & \omega<0
\end{cases}
\]
then 
\begin{eqnarray*}
\Psi^{c}(\omega) & = & {\cal F}(\Psi^{c}(x))\\
 & = & {\cal F}(\Psi_{h}(x)+j\Psi_{g}(x))\\
 & = & \Psi_{h}(\omega)+j\Psi_{g}(\omega)
\end{eqnarray*}
and 
\[
\Psi^{c}(\omega)=\begin{cases}
2\Psi_{h}(\omega) & \omega>0\\
0 & \omega<0
\end{cases}
\]

Thus, the Fourier transform of complex wavelet transform $\Psi^{c}(\omega)$ has no energy in the negative frequency region. It makes DTCWT an analytic wavelet transform with analytic output signals. Due to this analyticity, the dual-tree wavelet transform has implicitly managed to include all information in the half positive plan of the frequency domain.


It is quite straight forward for 2-D DWT expansion from 1-D DWT, discussion in the previous subsection \ref{subsec:dwt}. However, it is unfortunately not easy for similar expansion from 1-D DTCWT to 2-D DTCWT transforms because Hilbert Transform (HT) and analytic signals need an theoretical extension for 2-D signals. Furthermore, there exists not only one but several definitions which define different zero-out regions ( negative frequency domain in 1D DT-CWT) , signal-power regions (positive frequency domain in 2D DT-CWT). In this paper, we only focus on Bulow definition \cite{Bulow2001} of analytic quaternion signals which combines both partial and total Hilbert transform (HT) . Partial HTs are done along either $x$ or $y$ directions only; meanwhile, total HT is carried out on both directions simultaneously. They are defined as following formula.

\begin{eqnarray*}
f_{{\cal H}_{i_{1}}}(\mathbf{x}) & = & f(\mathbf{x})\circ\frac{\delta(y)}{\pi x}\\
f_{{\cal H}_{i_{2}}}(\mathbf{x}) & = & f(\mathbf{x})\circ\frac{\delta(y)}{\pi x}\\
f_{{\cal H}_{i}}(\mathbf{x}) & = & f(\mathbf{x})\circ\frac{1}{\pi^{2}xy}
\end{eqnarray*}

\begin{figure}[!htbp]
\centering
\includegraphics[width=0.75\textwidth]{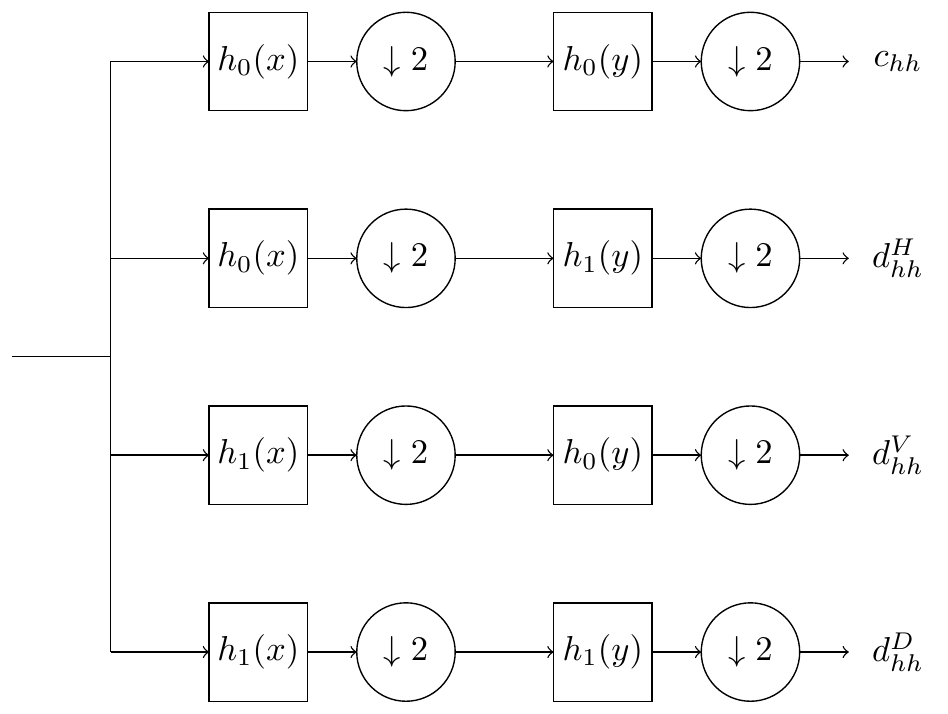}
\caption{Illustration of 2D dual-tree complex wavelet transform}
\label{fig:dtcwt2d_illus}
\end{figure}

The $f_{{\cal H}_{i_{1}}},f_{{\cal H}_{i_{2}}}(\mathbf{x})$ are partially Hilbert transformed along $x$ and $y$ axis consequently, and $f_{{\cal H}_{i}}(\mathbf{x})$ is total HT; while $\circ$ denotes 2-D convolution. Each 2-D CWT basis is a complex analytic function, computationally equivalent to a product of two 1-D complex wavelet functions either along only one or both axis. Similar to
expansion of discrete real wavelet, the diagonal sub-band wavelet is defined as $f(\mathbf{x})=\psi_{h}(x)\psi_{h}(y)$. Other total and partial HT are products of coefficients from different sets of
wavelet functions deployed in the 1-D CWT implementation. 
\[
(f_{{\cal H}_{i_{1}}},f_{{\cal H}_{i2}},f_{{\cal H}_{i}})=(\psi_{g}(x)\psi_{h}(y),\psi_{h}(x)\psi_{g}(y),\psi_{g}(x)\psi_{g}(y))
\]
To unify all different Hilbert Transform in a meaningful and compact representation, we can utilize quaternion algebra and treat  $f(\mathbf{x})$ as a real part and $(f_{{\cal H}_{i_{1}}},f_{{\cal H}_{i2}},f_{{\cal H}_{i}})$ as three imaginary components \cite{Chan2008}.
\[
f_{A}^{q}(\mathbf{x})=f(\mathbf{x})+j_{1}f_{{\cal H}_{i_{1}}}(\mathbf{x})+j_{2}f_{{\cal H}_{i_{2}}}(\mathbf{x})+j_{3}f_{{\cal H}}(\mathbf{x})
\]
More details about theory behind QWT and its special characteristics such as its singular cases, three phases, and zero-out regions can be found in Chan etal 's and Bulow 's publications \cite{Chan2008,Bulow2001}. Resting on form of the above quaternion wavelet transformation, we can organize four quadrant components of 2-D wavelet $(f\,,\, f_{{\cal H}_{i_{1}}},f_{{\cal H}_{i2}},f_{{\cal H}_{i}})$ as a quaternion. Lets take a example of diagonal signals with following quadrant components.
\[
(f\,,\, f_{{\cal H}_{i_{1}}},f_{{\cal H}_{i2}},f_{{\cal H}_{i}})=(\psi_{h}(x)\psi_{h}(y),\psi_{g}(x)\psi_{h}(y),\psi_{h}(x)\psi_{g}(y),\psi_{g}(y)\psi_{g}(y))
\]
We can have a diagonal quaternion wavelet functions for the diagonal sub-band mathematically defined as follows.
\[
\psi^{D}(x,y)=\psi_{h}(x)\psi_{h}(y)+j_{1}\psi_{g}(x)\psi_{h}(y)+j_{2}\psi_{h}(x)\psi_{g}(y)+j_{3}\psi_{g}(x)\psi_{g}(y)
\]
To compute the QWT coefficients, we can use proposals of a separable 2-D implementation \cite{Kingsbury2001} of dual-tree filter-banks previously illustrated in the figure \ref{fig:dtcwt_illus}. At each filtering stage, both two-sets of wavelet filters $h$ and $g$ are independently applied to each dimension $x$ and $y$ of a 2-D image. For example, the filter-bank $h$ is applied along both axis; then, it yields the scaling coefficients $c_{hh_{S}}$ and three diagonal,vertical and horizontal wavelet coefficients $d_{hh_{s}}^{D},d_{hh_{s}}^{V}$ and $d_{hh_{s}}^{H}$ respectively as shown in the figure \ref{fig:dtcwt2d_illus}.
Dual-tree implementation of two separated filter-banks for 1-D signal can be considered as four independent filter banks for 2-D signals according to all possible combinations of filter for one dimension $(hh,hg,gh,gg)$. With these combinations of filters and corresponding wavelet functions $\psi(x)\phi(y),\phi(x)\psi(y)$ and $\psi(x)\psi(y)$ are generated four components of quaternion wavelet transform for horizontal, vertical, and diagonal sub-bands. Four different wavelet coefficients from these filter banks are arranged by quaternion algebra to obtain QWT coefficients. For example, a coefficient from diagonal wavelet sub-band of QWT can be written in terms of responses from independent filter-banks as follows.
\[
d_{s}^{D}=d_{hh}^{D}+j_{1}d_{gh}^{D}+j_{2}d_{hg}^{D}+j_{3}d_{gg}^{D}
\]
So far, we have taken a diagonal sub-band as example for showing how QWT can be computed. The construction and properties for other two sub-bands are similar to what have been done for diagonal sub-bands. Except that the axis combinations results in a horizontal sub-band $\psi(x)\phi(y)$ or for a vertical sub-band $\psi(x)\psi(y)$ instead of a diagonal sub-band $\psi(x)\psi(y)$. In summary, QWT at each stage sports three quaternion sets corresponding to three sub-bands; each quaternion contains four wavelet functions. Therefore, there are 12 functions in total which can be easily seen as matrix of functions as follows. 
\begin{eqnarray*}
\left[\begin{array}{ccc}
d_{s}^{H} & d_{s}^{V} & d_{s}^{D}\end{array}\right] & \overset{q}{=} & \left[\begin{array}{ccc}
d_{hh}^{H} & \quad d_{hh}^{V} & \quad d_{hh}^{D}\\
d_{gh}^{H} & \quad d_{gh}^{V} & \quad d_{gh}^{D}\\
d_{hg}^{H} & \quad d_{hg}^{V} & \quad d_{hg}^{D}\\
d_{gg}^{H} & \quad d_{gg}^{V} & \quad d_{gg}^{D}
\end{array}\right]\\
 & = & \left[\begin{array}{ccc}
\psi_{h}(x)\phi_{h}(y) & \phi_{h}(x)\psi_{h}(y) & \psi_{h}(x)\psi_{h}(y)\\
\psi_{g}(x)\phi_{h}(y) & \phi_{g}(x)\psi_{h}(y) & \psi_{g}(x)\psi_{h}(y)\\
\psi_{h}(x)\phi_{g}(y) & \phi_{h}(x)\psi_{g}(y) & \psi_{h}(x)\psi_{g}(y)\\
\psi_{g}(y)\phi_{g}(y) & \phi_{g}(y)\psi_{g}(y) & \psi_{g}(y)\psi_{g}(y)
\end{array}\right]
\end{eqnarray*}
Columns of the above matrix correspond to quaternion wavelet functions of the horizontal sub-band $d^H$, the vertical sub-band $d^V$, and diagonal sub-band $d^D$ from left-to-right respectively. The three according wavelet coefficients are $d_{s}^{H}$, $d_{s}^{V}$ and $d_{s}^{D}$ and the $\overset{q}{=}$ operator means formation of quaternion number by coefficients along each column.
Though quaternion wavelet coefficients possess rich phase information, our research currently focuses on magnitudes of each wavelet sub-bands. Therefore, magnitudes of horizontal, vertical sub-bands can be computed according to quaternion magnitude formula as follows.
\begin{eqnarray*}
\Vert d_{s}^{H}\Vert & = & \sqrt{(d_{hh}^{H}).^{2}+(d_{gh}^{H})^{2}+(d_{hg}^{H}).^{2}+(d_{gg}^{H})^{2}}\\
\Vert d_{s}^{V}\Vert & = & \sqrt{(d_{hh}^{V}).^{2}+(d_{gh}^{V})^{2}+(d_{hg}^{V}).^{2}+(d_{gg}^{V})^{2}}\\
\Vert d_{s}^{D}\Vert & = & \sqrt{(d_{hh}^{D}).^{2}+(d_{gh}^{D})^{2}+(d_{hg}^{D}).^{2}+(d_{gg}^{D})^{2}}
\end{eqnarray*}
While the final approximated version of input signals, which are not decomposed further by the transform, have its magnitude computed by quaternion algebra.
\begin{eqnarray*}
\Vert c_{S}\Vert & = & \sqrt{(c_{hh}).^{2}+(c_{gh})^{2}+(c_{hg}).^{2}+(c_{gg})^{2}}\\
 & = & \sqrt{(\phi_{h}(x)\phi_{h}(y)).^{2}+(\phi_{g}(x)\phi_{h}(y))^{2}+(\phi_{h}(x)\phi_{g}(y)).^{2}+(\phi_{g}(y)\phi_{g}(y))^{2}}
\end{eqnarray*}

\subsection{Quaternion Wavelet Packet Transform}
\label{subsubsec:qwptbb}

To construct a packet form of QWT, each and every sub-band $c_{S},d^{H},d^{V},d^{D}$ should be repeatedly decomposed by low-pass $(h_{0,}g_{0})$ and high-pass filters $(h_{1},g_{1})$. Bayram \etal \cite{Bayram2008} has investigated into formation of wavelet packets for DT-CWT, an equivalent form of QWT. In order to get an analytic quaternion wavelet packet, the filter banks need to be chosen in a specific way such that the Hilbert transform relationship is preserved. In Bayram's works \cite{Bayram2008}, the analytic wavelet transformed can be achieved if whatever filter-bank is used to decompose the first filter-bank of QWT should also be used for the second (dual ) filter-banks. Another important point about the extension to wavelet packet QWPT is the choice of the extension filters $f_{i}(x).$ It has been  found that the only necessary constrain to preserver the Hilbert transform property is forcing the usage of the same filter-pairs $f_{0}(x),f_{1}(x)$ in both filter-banks of QWT or DT-CWT . Therefore, any CQF pair of filter-banks with short support, frequency selectivity or possessing a number of vanish moments can be candidates for the extension filter. Noted that, the above criteria such as CQF pair of filters have been employed for extending a regular DWT. Like other derivatives of DT-CWT or QWT, the quaternion wavelet packet transform (QWPT) are approximately shift-invariant, which means the energy
in each sub-band is approximately preserved if the input signals are shifted by a number of samples. Noteworthy that, there are other methods beside QWPT with shift-invariant property in wavelet pack decomposition. For example, by performing an exhaustive search over all shifted wavelet packet bases to find the ``best basis'' according to a certain cost function \cite{Cohen1997}, the orthonormal wavelet packet transform becomes shift-invariant in a sense that energy in each sub-band is invariant to transition of input signals . This (approximately) shift-invariance property becomes very useful and important in the search for a suitable energy descriptors. This shift-invariance property guarantees that DT-CWT, DT-CWPT or QWT, QWPT would have energy descriptors robust to certain amount  of affine transformation in input signals. 

Interactions of filtering both low and high components at each stage of DT-CWT introduces a complete structures of all possible sub-bands that can be generated by the filter-bank pair. Each tree forms a
unique frequency profile of input signals. Among those countless numbers of possibilities, there exist a frequency decomposition being more sparse and compact than the others. It is called the best-basis in
terms of representing the input signals with fewest wavelet coefficients. A fast algorithm for indicating such best basis has been reported in extension from DWT to DWPT \cite{Coifman1992}. In addition, it is previously mentioned in the section \ref{subsubsec:dwptbb}, the same strategy can be adopted for searching best-basis in QWT. In brief, the approach is looking for a path in a binary of decomposition
to minimize a Shannon entropy cost function; more details can be found in the work of Coifman \etal \cite{Coifman1992}. After ``best basis'' searching for QWPT decomposition, we can identify magnitudes and energy of coefficients at a specific location by a simple quaternion algebra. $\Vert q(a,b,c,d)\Vert=\sqrt{a^{2}+b^{2}+c^{2}+d^{2}}$.

\section{Wavelet Coefficients Correlation}
\label{sec:distribution2d}
\begin{figure}[!htbp]
	\centering
		\includegraphics[width=\textwidth]{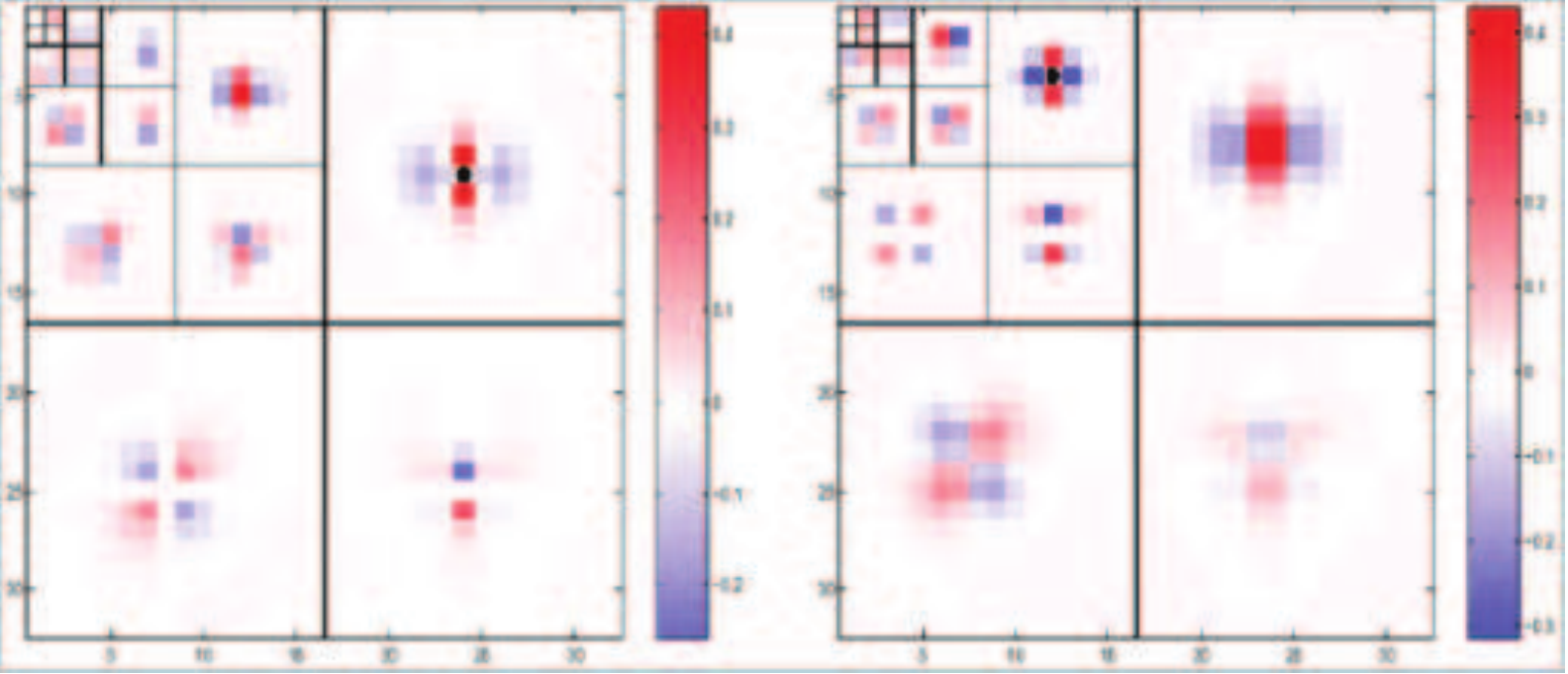}
	\caption{Illustration of wavelet coefficients inter-band and intra-band correlation \cite{Azimifar2011}}
	\label{fig:wavelet_correlations}
\end{figure}

The previous section \ref{sec:distribution1d} have discussed the potential of using energy density distribution of localized time-scale element or wavelet elements instead of pixel-value probability distribution. Only general 1-D signal is considered and these elements are assumed to be independent or at least linearly independent (uncorrelated); however, this assumption only works for random variables as input signals. Practically, except total noise, any meaningful signals often has specific structures persistent across multiple time-scale element in 1-D case. For 2-D signals like natural images, an additional orientation needs considering; in other words, their wavelet coeffcients are highly statistically related across scales, orientation, spaces. This phenomenon is systematically studied and confirmed in Azimifar \etal research \cite{Azimifar2011}. The author has conducted an empirical study of joint wavelet statistics for texture and natural images to investigate correlation relationship between neighbouring coefficients. Examination of these dependencies helps propose appropriate models for such a transform-domain algorithm. Though Azimifar's work \cite{Azimifar2011} only covers linear dependencies and just a squint on non-linear relations, its proposals are evaluated on a collection of 5000 real images. Therefore, we believe her conclusion in that study is generally true at least for natural images, the main researching objects. In brief, there exists a few elementary correlation relationships as follows.
\begin{itemize}
\item The spatially-localized and sparse correlation structure has a clear persistence across scales.
\item Every coefficient exhibits correlations extending across multiple scales, with spatially near neighbors both within and across orientations.
\item A subband coefficients at the same spatial locations but from different orientations are not linearly correlated.
\item Within-subband, inter-subband, and inter-scale correlations are highly oriented and persistent across local neighbors of its parent.
\end{itemize}
The below figure \ref{fig:wavelet_correlations} clearly illustrates all mentioned correlations, their preferences to locality as must be expected. This locality increases toward finer scales, which supports persistency property of wavelet coefficients. A single coefficient correlates with its parents as well as neighbors across orientations and scales. 
Among several mentioned statistical dependencies, the most vital findings for our work are uncorrelated siblings coefficients across orientation and strong correlated coefficients across scales since it theoretically allows uncertainty and mutual information estimation of a 2D time-scale element or a wavelet sub-band energy descriptor. To elaborate this point, lets consider two adjacent scales $s1,s2$ and their corresponding coefficients $w_{i}f(x,y,s_1)$ of horizontal, vertical, diagonal orientation $i={v,h,d}$ for 2-D signals $f(x,y)$. Due to non-correlation of sibling coefficients across orientation, it is possible to consider three wavelet coefficients as a multivariate variable $W_s = (w_h,w_d,w_v)$ with uncertainty estimation by energy density distribution across three orientations $H(W_s)$. For two adjacent sub-band $s_1,s_2$, there are two multivariate variables $W_{s_1}$ and $W_{s_2}$ corresponding two entropy values $H(W_{s_1})$ and $H(W_{s_2})$, and the mutual information between two variables due to inter-scale dependencies between correspondent wavelet coefficients are computed as follows.
\begin{equation}
I(W_{s_1},W_{s_2}) = H(W_{s_1},W_{s_2})-H(W_{s_1})-H(W_{s_2})
\label{eq:mutualInfoExample}
\end{equation}
From that basic observation about inter-scale and intra-scale wavelet coefficients of natural images is developed the core idea of our proposal. More details about sub-band energy descriptors and how to measure their uncertainty and mutual information will be clearly explained in following sections \ref{subsec:descriptors}, \ref{subsec:intraDescriptor}

\subsection{Interscale Subband Energy Descriptor}
\label{subsec:descriptors}

Interesting relationship between basis-project methods and scale saliency are repeatedly discussed in several publications\cite{Kadir2001}, \cite{Kadir2003}. Kadir \cite{Kadir2001} actually discusses about behaviours of non-saliency and saliency regions in spectral and wavelet domain. A simple, flat, non-salient regions or images is sufficiently described by a single sub-band; meanwhile, complicated data and structure regions require more sub-bands descriptors. This directly introduces basis-projected sub-bands as potential alternative descriptors. Like pixel-value descriptors, real wavelet sub-bands must be treated as discrete variables due to its theoretical restriction, data-analysis uncertainties, $\sigma_{t}\sigma_{\omega} \geq \frac{1}{2}$. In other words, it is impossible for continuous wavelet sub-bands distribution at any specific location. Following available mathematical definition of PSS for discrete pixel descriptors, we sketch rough mathematical models of WSS with discrete sub-band descriptors, ${\lbrace e \in \mathbf{E}, \mathbf{E} = \lbrace e_1, e_2, \ldots e_m \rbrace \rbrace}$, in the equations \ref{eqnarr2:sl}, \ref{eqnarr2:hd}, \ref{eqnarr2:wd}, \ref{eqnarr2:sp} whereof $e$, $E$ are a element and set of sub-band descriptor consecutively. 
\begin{eqnarray} 
	Y_D(\vec{s_p},\vec{x}) &{}\triangleq{}& H_D(\vec{s_p},\vec{x}) W_D(\vec{s_p},\vec{x}) \label{eqnarr2:sl} \\
	H_D(s,\vec{x}) &{}\triangleq{}& -\sum_{d \in D} p_{b,s,\vec{x}}log_2{p(d,s,\vec{x})} \label{eqnarr2:hd}\\ 
	W_D(s,\vec{x}) &{}\triangleq{}& \frac{s^2}{2s-1} \sum_{d \in D} \vert p_{b,s,\vec{x}} - p_{b,s-1,\vec{x}} \vert \label{eqnarr2:wd} \\ 
	\vec{s_p} &{}\triangleq{}& \lbrace s : H_D(s-1,\vec{x}) < H_D(s,\vec{x}) >  H_D(s+1,\vec{x}) \rbrace \label{eqnarr2:sp} 
\end{eqnarray}
However, a general concept of sub-band descriptor is not useful in actual computation; therefore, an appropriate numerical attribute of sub-bands need proposing instead. Lets consider 2-D discrete real-wavelet transform with three sub-bands vertical ($v$), horizontal ($h$) and diagonal ($d$) sub-bands at each particular dyadic scale $s$ represented by three set of wavelet coefficients ($\mathbf{w_i}$) accordingly in the equation \ref{eqnarr3:wcf}. Equation \ref{eqnarr3:wed} uses those coefficients to compute sub-band energy densities as descriptors ($e$) for Wavelet-domain Scale Saliency (WSS).
\begin{alignat}{2}
	\mathbf{w_i} f(x,y,s_j) |_{ i=\lbrace v, h,d \rbrace,s_j=\lbrace s_1, s_2, ..., s_n \rbrace } &= f \left(x,y\right)*\psi_{s,i}\left(x,y,s_j\right)  \label{eqnarr3:wcf} \\
	\mathbf{P \lbrace w_i} f(x,y,s_j) \rbrace |_{i=\lbrace v,h,d \rbrace,s_j=\lbrace s_1, s_2, ..., s_n \rbrace} &= \vert \mathbf{w} f(x,y,s_j) \vert^2 \label{eqnarr3:wed}
\end{alignat}
In the standard real discrete wavelet transform (DWT), there are fixed three analysed sub-bands for each dyadic sampling step. Supposedly the maximum level of wavelet decomposition is $n$, the number of dyadic scales is $n$ with 3 sub-bands for each scale. With 4 or 5 as the usual number of decomposition levels, totally around 12 or 15 sub-bands descriptors are analysed for an image. This number of descriptors is significantly less than 255 pixel-value descriptors of PSS for any grey-scale image.

Besides wavelet transforms, different other types of basis projection techniques could also be utilized; for example, best basis wavelet packet analysis (DWPTBB). The full wavelet packet transform breaks signals into sub-bands with the same bandwidth at the maximum dyadic scale. It would not fit into the scale saliency concept which requires descriptors at different scales. Fortunately, the "balanced" full wavelet packet tree usually over-describes image properties, and the description can be optimized by Best Basis ($B^2$) finding operation. The optimized wavelet packet tree often has projected basis across dyadic scales since some small image details are best described with a basis at finer resolution while other big details prefer another basis with coarser resolution. The DWPTBB coefficients are utilized for sub-band energy density , the proposed image descriptors, calculation in the equations \ref{eqnarr4:wcf}, \ref{eqnarr4:wed}.
\begin{alignat}{10}
	\mathbf{w_i} f(x,y,s_j) |_{ (i,j) = \mathit{B^2} \left( \mathbf{w_i}f(x,y,s_j) \right)} &= f \left(x,y\right)*\psi_{s,i}\left(x,y,s_j\right)  \label{eqnarr4:wcf} \\
	\mathbf{P} \lbrace \mathbf{w_i} f(x,y,s_j) \rbrace |_{(i,j) = \mathit{B^2} \left( \mathbf{w_i}f(x,y,s_j) \right) } &= \vert \mathbf{w_i} f(x,y,s_j) \vert^2  \label{eqnarr4:wed}
\end{alignat}
Comparing mathematical statements \ref{eqnarr3:wcf},\ref{eqnarr3:wed} and \ref{eqnarr4:wcf},\ref{eqnarr4:wed} for sub-bands descriptors of Discrete Wavelet Transform (DWT) and Discrete Wavelet Packet Transform Best Basis (DWPTBB) consecutively, we can see their fundamental differences. While DWT provides determinant basis-projection methods with pre-computed basis and fixed structure of sub-bands, DWPTBB adapts itself into each data set. Then, its number and structure of sub-bands are specified by Best Basis ($B^2$) finding operator  \cite{Coifman1992}. It requires more operations; however, more faithful and adaptive descriptors can be achieved.

Both DWTBB and DWT are popular wavelet-transforms; however, they both depend on shift-variant real discrete wavelet transforms. It means that projection of coefficients not only depends on data but also its relative location on the scene.
\begin{alignat}{5}
	\mathbf{w_i} f(x,y,s_j) &\neq \mathbf{w_i} f(x+\Delta(x),y+\Delta(y),s_j) & &, \exists x,y,s_j,\mathbf{w_i},\Delta(x),\Delta(y) \\
	P \lbrace \mathbf{w_i} f(x,y,s_j) \rbrace &\neq P \lbrace \mathbf{w_i} f(x+\Delta(x),y+\Delta(y),s_j) \rbrace & &, \exists x,y,s_j,\mathbf{w_i},\Delta(x),\Delta(y)	
\end{alignat}
As the fourth criteria for good information measurement of Starck \etal \cite{Starck1999} states that entropy must work in the same way regardless of descriptors' locations. Both DWT and DWTBB projected descriptors do not satisfy that condition since usages of these descriptors might lead to different information estimation for identical data at two different locations. The shift-variance of real-wavelet transform can be avoided by complex wavelet transform design; for instances, recently developed dual-tree complex wavelet transform (DTCWT) \cite{Selesnick2005}, Quaternion wavelet transform (QWT) \cite{Chan2008}, or dual-tree complex wavelet packet transform with best-basis (DTCWTBB) \cite{Bayram2008}. General formula of complex coefficients and their corresponding sub-band energy density are summarized in the equations \ref{eqnarr5:wcf},\ref{eqnarr5:wed}.
\begin{alignat}{2}
	& \mathbf{w_i} f(x,y,s_j) |_{i  =\lbrace \lbrace v,h,d \rbrace \vee B^2 \left( \mathbf{w_i} f \right) \rbrace} = f \left(x,y\right)* \left( \psi_{g,s,i}\left(x,y,s_j\right)  + j\psi_{h,s,i}\left(x,y,s_j\right) \right) \label{eqnarr5:wcf} \\
	& \mathbf{P} \lbrace \mathbf{w_i} f(x,y,s_j) \rbrace |_{ s_j  =\lbrace s_1,... ,s_n \rbrace \vee  B^2 \left( \mathbf{w_i} f \right) } = \Vert \mathbf{w_i} f(x,y,s_j) \Vert_2^2 \label{eqnarr5:wed}
\end{alignat}
Dual-tree approaches use two different wavelet filter-banks ,$\lbrace \psi_g, \psi_h \rbrace$, and they are designed to form analytical complex filter banks, $\lbrace \psi_g(x,y) + j \psi_h(x,y), \psi_h(x,y) \approx H \left( \psi_g(x,y) \right) \rbrace$. The magnitudes of projected-complex coefficients are proven to be shift-invariant; therefore, its derived energy density of the sub-bands is as well shift-invariant. Probably, the quaternion version of wavelet transform (QWT) and quaternion wavelet packet transform best basis (QWTBB) with shift-invariant property would provide better descriptors than their real counterparts according to five criteria of Starck \cite{Starck1999}.

\subsection{Intra-scale Subband Energy Descriptor}
\label{subsec:intraDescriptor}

As previously mentioned in the section \ref{sec:distribution2d}, there is strong correlation or statistic linear dependence between wavelet coefficients in natural images. The first correlation, the inter-scale dependencies, has been discussed in the section \ref{sec:distribution1d},\ref{sec:distribution2d}, and modeled as sub-band descriptors in the previous section \ref{subsec:descriptors}. Moreover, the relation have been widely and effectively employed in various tree-structured coding techniques such as SPIHT \cite{Buccigrossi1999}. Besides inter-scale relationship, many authors \cite{Do2005} have pointed out another strong correlation of intra-band coefficients existing across many different types of natural scenes . Minh Do and M Vetterli  \cite{Do2005} successfully modelled coefficients of a wavelet sub-band with a simple explicit mathematical form, Generalized Gaussian Distribution (GGD). While statistical distribution of wavelet coefficients gets a lot of interest, several researches have proposed different mathematical models for analysing this statistical characteristic. However, few models of wavelet coefficients marginal density at a particular sub-band works better than GGD in terms of accuracy, approximation and simplicity. After such the distribution is widely observed in experimental data with natural images, theoretical analysis on the plausibility of modelling by the GGD distribution is defined as follows. 
\[
p(x;\alpha,\beta)=\frac{\beta}{2\alpha\Gamma(1/\beta)}e^{\left(-\vert x\vert/\alpha\right)^{\beta}}
\]
where $\Gamma(z)=\int_{0}^{\infty}e^{-t}t^{z-1}dt,\, z>0$ is the Gamma distribution. Here $\alpha$ dictates the scale parameter or variance of the distribution, and $\beta$ controls shapes. For example, GGD with$\beta=1$ is Gaussian distribution; it becomes Laplacian distribution with $\beta=2$.

\section{Information Measurement}
\label{sec:calculation}
In the previous section \ref{subsec:descriptors}, four different wavelet transforms generate corresponding wavelet sub-band energy density descriptors. From those energy density, energy probability distribution function ($PDF$) at each scale $s_j$ can be computed as follows.
\begin{eqnarray*}
p_{interband}(x,y,s_{j}) & = & p\lbrace\mathbf{P}\left[\mathbf{w_{i}}f(x,y,s_{j})\right]\rbrace|_{i=\lbrace v,h,d\rbrace\vee i=B^{2}(\mathbf{w_{i}})\rbrace,j<=m}\\
 & = & \frac{\mathbf{P}\left[\mathbf{w_{i}}f(x,y,s_{j})\right]}{\sum_{j}\sum_{i}\mathbf{P}\left[\mathbf{w_{i}},f(x,y,s_{j})\right]}  \label{eqnarr6:pdf}
\end{eqnarray*}
The above formula computes the probability of energy density at one location $(x,y)$ across different sub-bands, $i = \lbrace v,h,d \rbrace $ for (DWT) or (QWT) or $ i = B^2(\mathbf{w_i})$, for (DWPTBB) and (QWPTBB) from the smallest scale, $1$, to currently considered scales, $m$. The first level uses the smallest sampling window size of wavelet atoms; therefore, it generates analysed coefficients with finest details. Then, the sampling window sizes are doubled after each level; they generate coarser analysed details. It is quite similar to PSS sampling operations except that scales are doubled rather than increased by a unit. Like $PDF$ of PSS descriptor, WSS descriptors $PDF$ are distributed with increasing scales of j, from level $1$ ( smallest wavelet atom ) to level $m$ ( currently biggest wavelet atom). From the equation \ref{eqnarr6:pdf}, it is straightforward to compute feature-space entropy $H_{Observer} \left(x,y,s_m \right)$ as follows whereof $p \lbrace \mathbf{P} \left[ \mathbf{w_i} f(x,y,s_j) \right] \rbrace$ is shorted as $p_{interband}(x,y,s_j)$ .
\begin{equation}
	-\sum_{\lbrace i = \lbrace v,h,d \rbrace \vee i = B^2(\mathbf{w_i}) \rbrace , \lbrace j <= m \rbrace } p_{interband}(x,y,s_j) \log p_{interband}(x,y,s_j) 
	\label{eqnarr7:ent}	
\end{equation}
Both entropy of PSS's descriptors and the above entropy formula for the proposed descriptor only summarizes statistical property in local spatial regions since both considering window sizes in PSS and scale levels of wavelet decomposition are finite. Then, it lacks involvement of energy distribution in the whole image and it is confirmed that such distribution is vital for natural image and texture modeling \cite{Do2005}. As presented in the sub-section \ref{subsec:intraDescriptor} is the Generalized Gaussian Distribution of coefficients magnitudes from a wavelet intra-band.

\begin{equation}
p_{intraband}(x,y,s_j,\alpha,\beta)=\frac{\beta}{2\alpha\Gamma(1/\beta)}e^{\left(- \sqrt{x^2+y^2} /\alpha\right)^{\beta}}
\label{eqnarr8:ggd}
\end{equation}

In order to combine both global and local characteristics into a single value, we propose cross-entropy $H_{Searcher}\left(x,y,s_m \right) $  between inter-band and intra-band distribution as an alternative formulation of the equation \ref{eqnarr7:ent}. 

\begin{equation}
	-\sum_{\lbrace i = \lbrace v,h,d \rbrace \vee i = B^2(\mathbf{w_i}) \rbrace , \lbrace j <= m \rbrace } p_{interband}(x,y,s_j) \log p_{intraband}(x,y,s_j) 
	\label{eqnarr9:crossent}	
\end{equation}

To distinguish between two modes of entropy computation, we names the local entropy by the equation \ref{eqnarr7:ent} as "observer" mode, and the cross-entropy involving both local and global statistics as "searcher" mode. In later formula, when general entropy symbol $H$ without specific subscripts appears in any formulas, it means both modes are eligible for those equations. Those names also help to distinguish different parameters and simulation modes presented in the experimental sections \ref{subsec:result1}.

The equation \ref{eqnarr7:ent} computes feature-space entropy of sub-band energy descriptors for WSS as the equation \ref{eqnarr2:hd} does for PSS. Half of scale saliency measure, feature-space entropy, has been figured out for sub-band energy density descriptors. The other half of the problem rests in computational details of inter-scale saliency; in other words, how the equation \ref{eqnarr2:wd} should be interpreted with the proposed descriptors. In equation \ref{eqnarr2:hd}, the inter-scale saliency is measured as total variation in probability distribution of descriptors at two consecutive scales in which pixel-value descriptors ($d$) appear in both distributions, it complicates the problem. However, the situation is different for wavelet sub-band energy density descriptors since each sub-band in the current level is unique for this level only. It does not appear in other levels of analysis. This wonderful property simplifies out task in building sub-band probability distribution for different levels but makes the equation \ref{eqnarr1:wd} inappropriate for sub-band features. Since it is unjustifiable to find total variation of two $PDF$ on two different set of descriptors, an alternative interpretation of inter-scale saliency need developing. Lets consider $ P(M) = \lbrace p_{i,j}(x,y,s_j) | \forall i, j <= m \rbrace$, $PDF$ of all sub-bands up to the current level, $m$. When a new analysed sub-band, $D = \lbrace p_{i,j}(x,y,s_j) | j = m + 1 \rbrace$, is generated, this sub-band descriptor will modify the current $PDF$ into $P(M \vert D)$. The distance between the prior model and the modified model can be measured by Kullback-Leibler divergence as follows.
\begin{equation}
	K( P(M|D),P(M) ) = \int_M P(M|D) \log \frac{P(M|D)}{P(M)} \label{eqnarr8:kld}
\end{equation}
Noteworthy, it is similar to Itti's Bayesian Surprise Saliency (BSS) metric \cite{Baldi2010}, and the surprise model can be extended for multiple sub-bands descriptors or evidences in BSS. The equation \ref{eqnarr8:kld} becomes mutual information between the current model and a set of new evidences. In other words, the expectation of surprise for adding new sub-bands into the current model is the mutual information between new sub-bands and the current model, shown in the equation \ref{eqnarr9:mif}. 
\begin{equation}
	MI(D,M) = \int_D K( P(M|D),P(M) ) = \int_{D,M} P(D,M) \log \frac{P(D,M)}{P(M)P(D)} \label{eqnarr9:mif}
\end{equation}
Therefore, mutual information is chosen as inter-scale saliency for successive dyadic scales since it actually implies averaged "bayesian surprise" \cite{Baldi2010} saliency of sub-bands across scales. Furthermore, mutual information as inter-scale saliency measurement well emphasizes the structural coherence of data across scales. If there are useful structures such as edges or joints and they are consistent across consecutive scales, they will increase mutual information between two consecutive scales. Otherwise noises have no mutual information across scales as its self-information is zero, $I(N,N) = 0$. It is remarkable that mutual information satisfies the fifth criterion of the good information estimation by Starck \etal \cite{Starck1999}. The only remaining step is identifying how the mutual should be calculated in discrete cases. Following formula shows relation between mutual information and entropy.
\begin{alignat}{2}
	MI(D,M) &=&\,& \quad H(D) \quad + \quad H(M) \quad - \quad H(D,M) \label{eqnarr10:mif} \\
	H(M) &=&& - \sum_{\lbrace \lbrace i = \lbrace v,h,d \rbrace \vee i = B^2(\mathbf{w_i}) \rbrace , \lbrace j \leq m \rbrace  \rbrace} p_i(x,y,s_j) \log p_i(x,y,s_j)  \label{eqnarr10:entM} \\
	H(D) &=&& - \sum_{\lbrace \lbrace i = \lbrace v,h,d \rbrace \vee i = B^2(\mathbf{w_i}) \rbrace , \lbrace j = m \rbrace  \rbrace } p_i(x,y,s_j) \log p_i(x,y,s_j) \label{eqnarr10:entD} \\
	H(D,M) &=&& - \sum_{ \lbrace \lbrace i = \lbrace v,h,d \rbrace \vee i = B^2(\mathbf{w_i}) \rbrace , \lbrace j \leq m+1 \rbrace  \rbrace} p_i(x,y,s_j) \log p_i(x,y,s_j) \label{eqnarr10:entMD}
\end{alignat}
The mutual information can be directly calculated as difference between separated ($H(D)+H(M)$) and joint ($H(D,M)$) entropy estimation of the current energy descriptors (the current model) and the next-level sub-bands, the equation \ref{eqnarr10:mif}. While the entropy elements $H(D), H(M), H(D,M)$ can be easily estimated by simple mathematical equations \ref{eqnarr10:entM},\ref{eqnarr10:entD},\ref{eqnarr10:entMD}. The joint entropy $H(D,M)$ can be reused as $H(M)$ for the next level inter-scale saliency estimation because of the sub-band descriptors uniqueness. The scale saliency principles on wavelet-domain sub-band energy descriptors are summarized in the equation \ref{eqnarr11:entY} as product of maximum feature-space saliency and inter-scale saliency, or product of mutual information between consecutive levels and maximum sub-band entropy.
\begin{align}
	& H(M(x,y,s_p)) = - \sum_{i = \lbrace v,h,d \rbrace \vee i = B^2(\mathbf{w_i}), j \leq m } p_i(x,y,s_p) \log p_i(x,y,s_p)  \nonumber \\ 
	& MI(D(x,y,s_p),M(x,y,s_p-1)) = H(D) + H(M) - H(D,M) \nonumber \\ 
	& \vec{s_p} {}\triangleq{} \lbrace s : H(M(s-1,x,y)) < H(M(s,x,y)) \wedge H(M(s,x,y)) >  H(M(s+1,x,y)) \rbrace \nonumber \\ 
	& Y(M(x,y,s_p)) = H(M(x,y,s_p)) * MI(D(x,y,s),M(x,y,s_p-1)) \label{eqnarr11:entY}
\end{align}
The characteristic scale $s_p$ is chosen to maximize information of the model $H(M(s,x,y))$. Lets imagine the case prior scale contains only noise meanwhile later scales actually contain useful structures of images. With bias of Shannon entropy toward noise, the characteristic scale fails to enclose any useful structure. To overcome this drawback, we propose alternative approach, \textbf{DIS} to differentiate from the original strategy $WSS$, in which $s_p$ is selected so as to maximize inter-scale saliency or average "Bayesian surprise". \textbf{DIS} principles can be summarized as follows.
\begin{align}
	& \vec{s_p} {}\triangleq{} \lbrace s : MI(D_{s-1},M_{s-2}) < MI(D_{s},M_{s-1}) \wedge MI(D_{s},M_{s-1}) >  MI(D_{s+1},M_{s}) \rbrace \nonumber
\end{align}
Experiments with \textbf{DIS} and \textbf{WSS} are carried out and simulations results are detailed in the next section in order to confirm effectiveness of the proposed strategy. 

\section{Discussion \& Results}
\label{sec:results}

The previous sections \ref{subsec:descriptors} and \ref{sec:calculation} have analysed theoretical advantages of \textbf{WSS} and its derivative \textbf{DIS}.
In addition, the subsection \ref{subsec:descriptors} present four descriptors based on different wavelet transforms: DWT, QWT, DWPTBB, and QWPTBB. Accordingly, we have several derivatives for the proposed method according to specific choices of scale section mechanisms and sub-band descriptor. To evaluate them against other saliency approaches, they are compared with PSS \cite{Kadir2001}, and the de-facto ITT model \cite{itti1998model}. The purpose of comparisons are not for claiming the best saliency method or racing toward the highest possible evaluating measurement; it just proves the rationale of the assumption that feature and structural complexity would be a good clues for human attention. The best evaluation measurement reported does not necessarily mean the best saliency maps since it much depends on choices of databases, elimination of experimental bias, performance of human test subjects, etc. Moreover, a standardized evaluation process in saliency map evaluation is far from being reached since several researchers choose different database and measurement methods or even create their own. In our research, we focus on the effectiveness of information measurement in visual attention; then, the most common processes and databases would be chosen to confirm generalization of the assumption.

In line of searches for informative clues , Bruce and Tsotsos \cite{Bruce2009a} database is certainly among the popularly used stimuli. However, only Bruce's database is certainly not enough due to limits in numbers and contents of stimuli. Then, Kootstra's database \cite{Kootstra2011} are chosen for extra testing samples and ground-truths. Two database with over 200 samples with ground-truths provided by more than 50 human subjects would help to confirm the generalization of our proposed framework to a certain extent. Similarly, only common evaluation approaches are deployed in our studies, and they can be categorized into either quantitative or qualitative methods. Quantitative relations between different saliency methods and human visual performance are shown by appropriate statistical methods (AUC,NSS) with eye-tracking data as ground-truths. Meanwhile, the qualitative results, visual comparisons of different saliency maps, gives a glimpse about performance for each individual sample. It also specifies imaging contexts where saliency methods give reasonable solution as well as situations where saliency maps are unreasonable to human perception.
\subsection{Databases of image stimuli}
\label{subsec:database}
The ground-truth and data for basic evaluations of visual saliency performance is got from eye-tracking experiments. Specially in Neil Bruce database, 120 different color images are observed in random orders while there are 4 seconds gap between the previous and the next stimuli. To ensure consistency and accuracy of the database, subjects are asked to seat 0.75 m in front of a 21 inch CRT monitor. Especially, human subjects have no further instructions for any actions or clues for what images appear next. Furthermore, image contents are varied from indoor to outdoor environments. Sometimes, there are clear interesting objects in the scene; while some scenes are really general without any particular interests in any subjects. A non-head mount eye tracking apparatus  extracts locations of eye-fixation while human test subjects look at sample images. Other setting-up parameters are intended for a general-scene based stimuli which are typically found in urban environments. Moreover, the same parameters are used for collecting data from 20 different subject over 120 testing samples. The following figure \ref{fig5:nbdbsamples} shows first eight images from the Neil Bruce's database. 
\begin{figure}[!htbp]
	\centering
	\includegraphics[width=\textwidth]{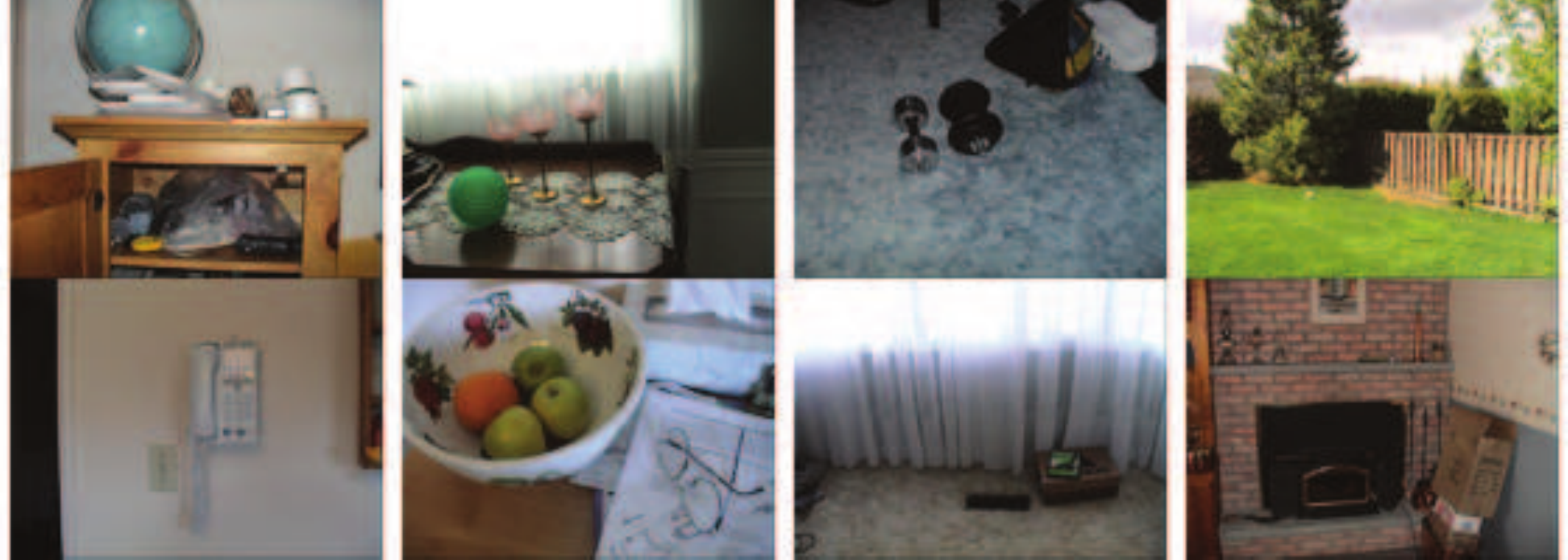}
	\caption{Neil Bruce's database}
	\label{fig5:nbdbsamples}
\end{figure}
Despite its popularity, images from Neil Bruce's database has narrow semantic content since it contains only urban scenes and mainly indoor environments. Besides that, the number of samples are relatively small. Due to that, an additional database should be included in simulations such that there more testing images of natural objects like animals, flowers, in natural environments. Kootstra's database \cite{Kootstra2011} have just satisfied these requirements with additional ground-truths for further experiments. Kootstra's ground-truths data are also collected from eye-tracking experiments although the experimental process is slightly different from what have been done to collect Bruce's database. In the psychological experiment, with head-mount eye-tracking devices, thirty-one students (15 men, 16 womens ) ranging from 17 to 32 of age took part in the experiments, and they are all naive about aims of experiments. Each human subject observes a total of 99 photographic image in five different categories while their eye movements are recorded simultaneously with the head-mounted device. There are nine-teen images in natural symmetry category; each of which contains symmetrical natural objects. Beside such symmetrical scenes, other non-symmetrical photographic scenes are included into the image sets such as: 12 images of animals in natural seeting, 12 images of street environments,  12 images of street scenes, 16 images of building and 40 images of natural environments. Figure \ref{fig6:ktdbsamples} gives an example of 5 categories of images in the Kootstra's database. Noted that,  each image is presented to viewers with a resolution of 1024x768 pixels on an 18" CRT monitor at a distance of 70 cm from the participant.
\begin{figure}[!htbp]
	\centering
	\includegraphics[width=\textwidth]{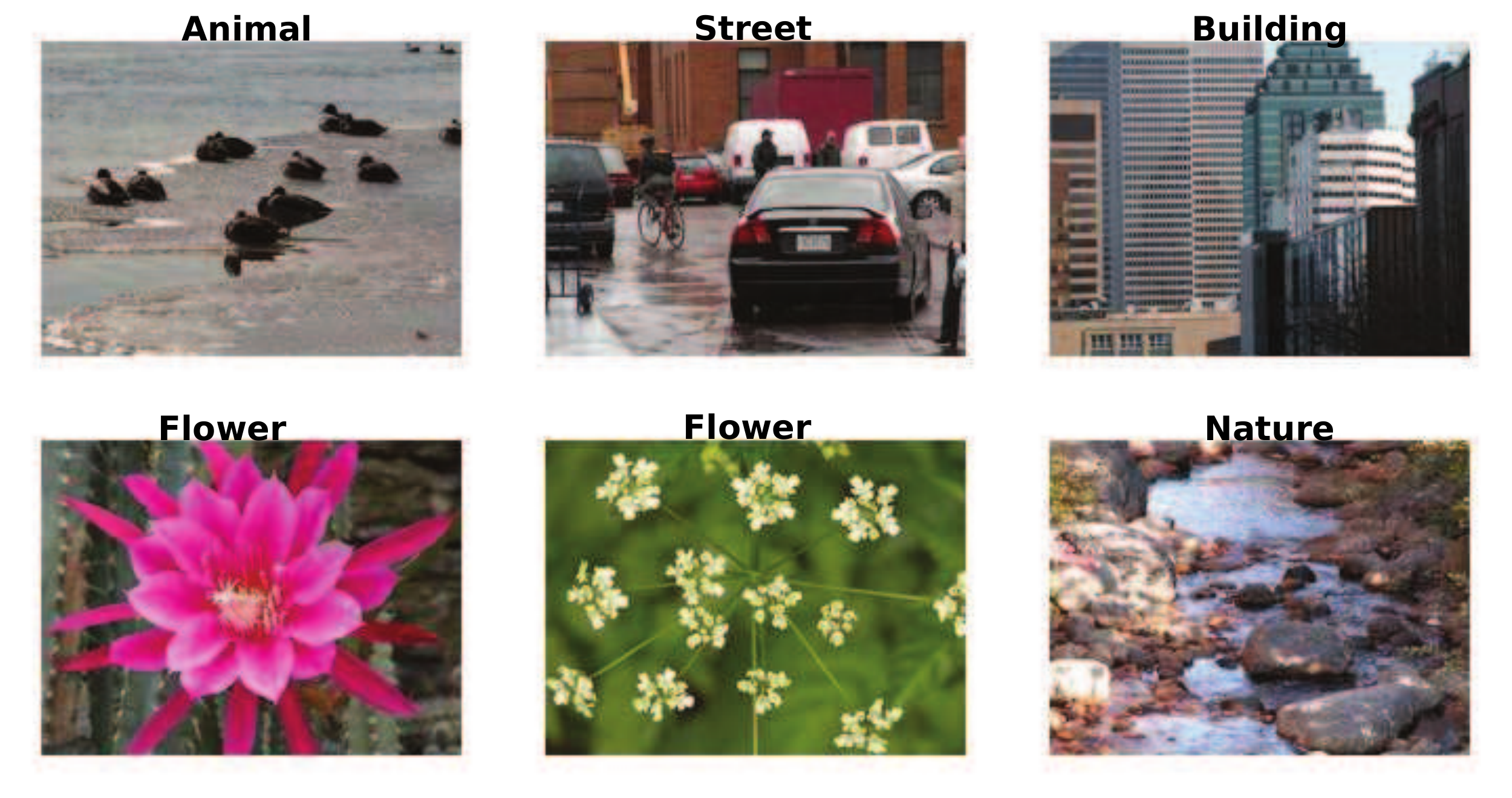}
	\caption{Kootstra's database}
	\label{fig6:ktdbsamples}
\end{figure}
\subsection{Quantitative Comparisons of Saliency Methods}
\label{subsec:result1}
The quantitative performance includes Receiver Operating Characteristics (ROC) curves with Area Under ROC Curve (AUC), and Normalized Scanpath Saliency (NSS) as numerical results. To ensure fair comparisons between methods, open-source evaluation codes for AUC and NSS \cite{Borji2012b} are employed. Noteworthy, saliency maps are standardized around median instead mean of distributions. Quantitative evaluation of visual saliency map on natural images with eye-tracking data ground has been initially studied by Tatler and recently summarized by Borji \etal \cite{Borji2012b}. More information about mechanisms behind ROC and AUC can be found \cite{Borji2012b}. In this section, we only focus on usages of these quantitative methods to compare, evaluate and prove rationale of our approach. As the main purpose of this evaluation is confirming effectiveness of informative clues in human's visual attention, our approach is not optimally tuned to reach the maximum AUC or NSS.

\begin{figure}[!htbp]
	\centering
 	\includegraphics[width=0.75\textwidth,natwidth=560,natheight=420]{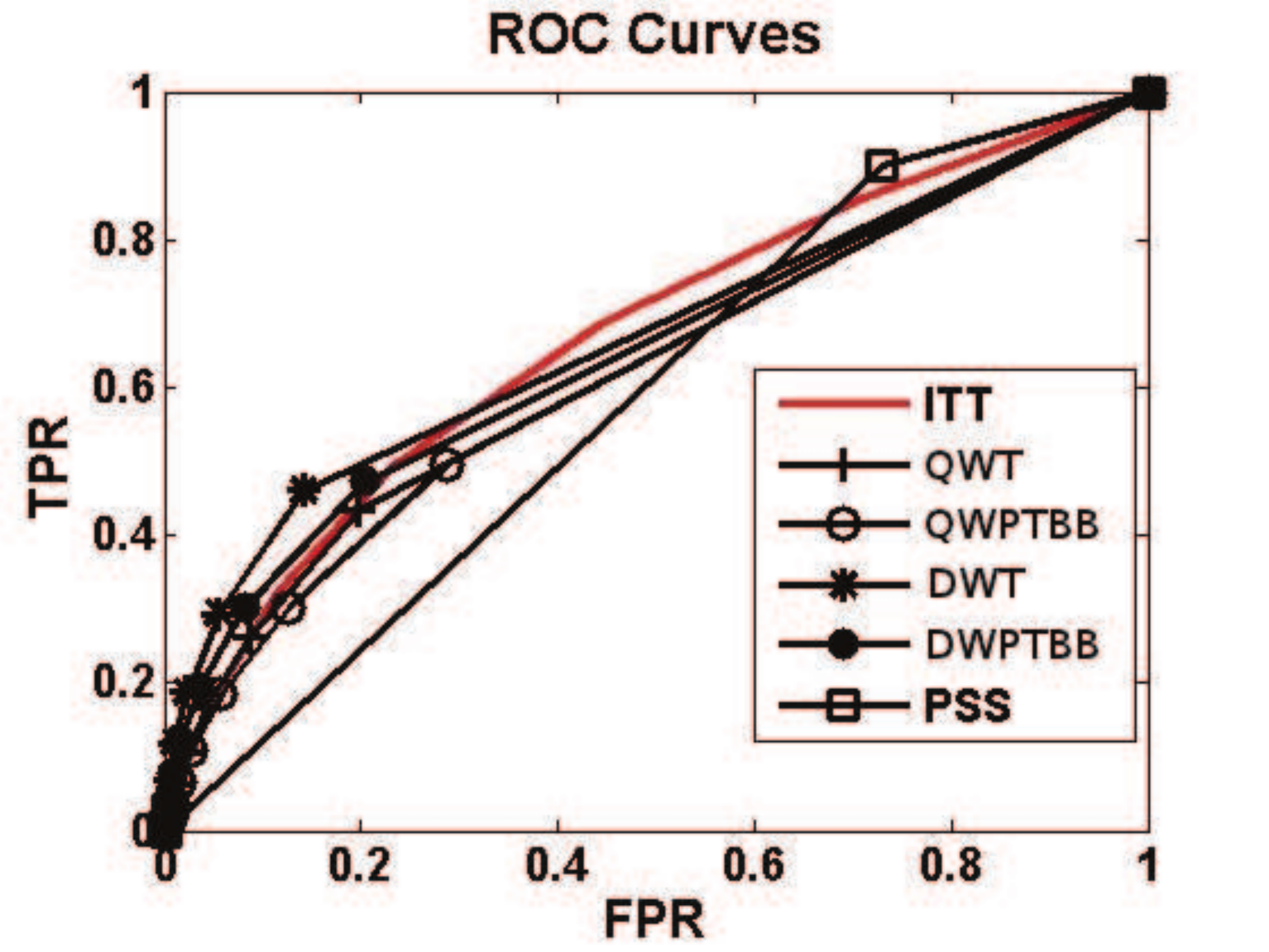}
\caption{ROC Curve - WSS}%
\label{fig1:roccuv}
\end{figure}

\begin{figure}[!htbp]
	\centering
	\includegraphics[width=0.75\textwidth,natwidth=560,natheight=420]{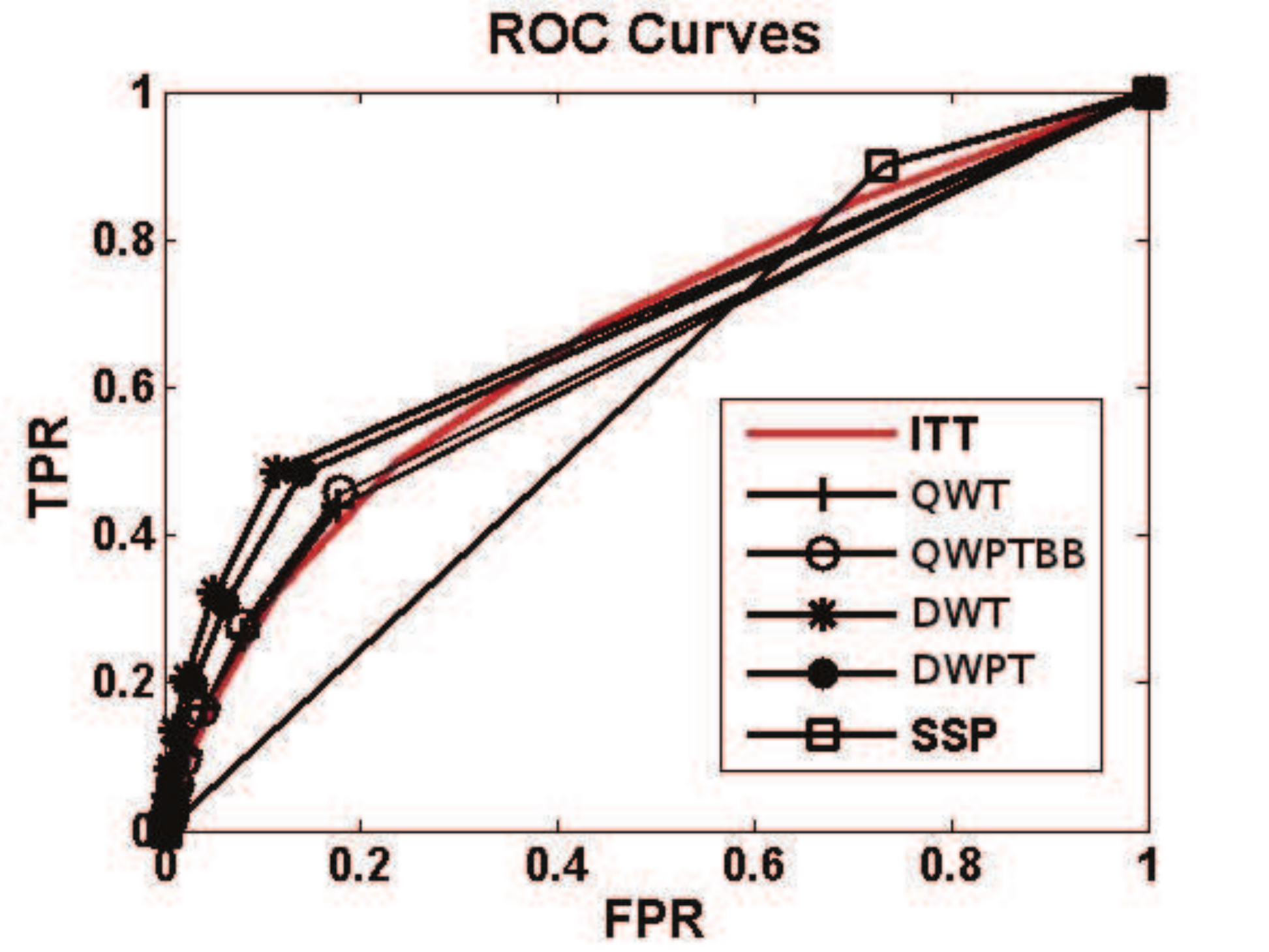}
	\caption{ROC Curve - DIS}%
	\label{fig2:roccuv}
\end{figure}

All four descriptors mentioned in the section \ref{subsec:descriptors} have been simulated with image samples from both Neil Bruce's and Kootstra's datasets. Noted that scale selection mechanisms have strong influences in formation of saliency maps; therefore, two separated simulations are carried out to investigate that effect as well. Figure \ref{fig1:roccuv} and \ref{fig2:roccuv} summarizes simulations results of proposed methods with corresponding WSS and DIS respectively in Neil Bruce's image dataset.
%

\begin{table}[!htbp]
	\centering
	\begin{tabular}{cccc}\hline
      		MTH & AUC & NSS & TIME(s) \\ \hline
       		ITT & 0.6944 & 0.27714 & 1.096s \\
		PSS & 0.5856 & -0.39175 & 7.1092s \\
        		DWT &  0.67823 & 0.33358 &  1.2401s \\
        		QWT & 0.66279 & 0.30002 & 1.9231s \\
        		DWPTBB &  0.6417 & 0.26079 & 2.6187s \\
        		QWPTBB & 0.63529 & 0.23714 & 5.2836s \\ \hline		
    	\end{tabular}     
	 \caption{Quantitative Result}%
	 \label{tab1:auctim}
\end{table}

According to the figure \ref{fig1:roccuv} and the table \ref{tab1:auctim}, performances of four WSS derivatives follow decreasing orders: DWT, QWT, DWPTBB and QWPTBB; however, all are better than PSS performance and comparable to ITT method. Especially, a computational time is deducted by approximately 7 times; noteworthy, the PSS is implemented in C++ with MATLAB interface and WSSs are totally written in MATLAB. For Niel Bruce database, the best basis approaches $DWPTBB$ and $QWPTBB$ does produce poorer results in both accuracy test, AUC and NSS as well as efficiency test, TIME.

\begin{table}[!htbp]
	\centering
	\begin{tabular}{cccc}\hline
      	MTH & AUC & NSS & TIME(s) \\ \hline
       	ITT & 0.6944 & 0.27714 & 1.096s \\
		PSS & 0.5856 & -0.39175 & 7.1092s \\
        DWT &  0.7028 & 0.3178 & 1.2689s \\
        QWT & 0.6922 & 0.3024 & 1.9527s \\
        DWPTBB &  0.6299 & 0.2546 & 2.4218s \\
        QWPTBB & 0.6394 & 0.2351 & 5.4835s \\ \hline	
    	\end{tabular}     
	 \caption{Quantitative Result}%
	 \label{tab2:auctim}
\end{table}

Quantitative performances of four \textbf{DIS} methods are shown in the figure \ref{fig2:roccuv} and the table \ref{tab2:auctim}. Mixed results are spotted. Performances of DWT and QWT descriptors with DSS approach are a little bit increased in terms of AUC if compared to the case of WSS. However, "Best-basis" descriptors (DWPTBB,QWPTBB) perform a little bit better if WSS are employed instead of DIS. Meanwhile, there is almost no difference between WSSs and DSSs in term of both NSS and TIME regardless descriptors.

\begin{figure}[!htbp]
	\centering
	\includegraphics[width=0.75\textwidth]{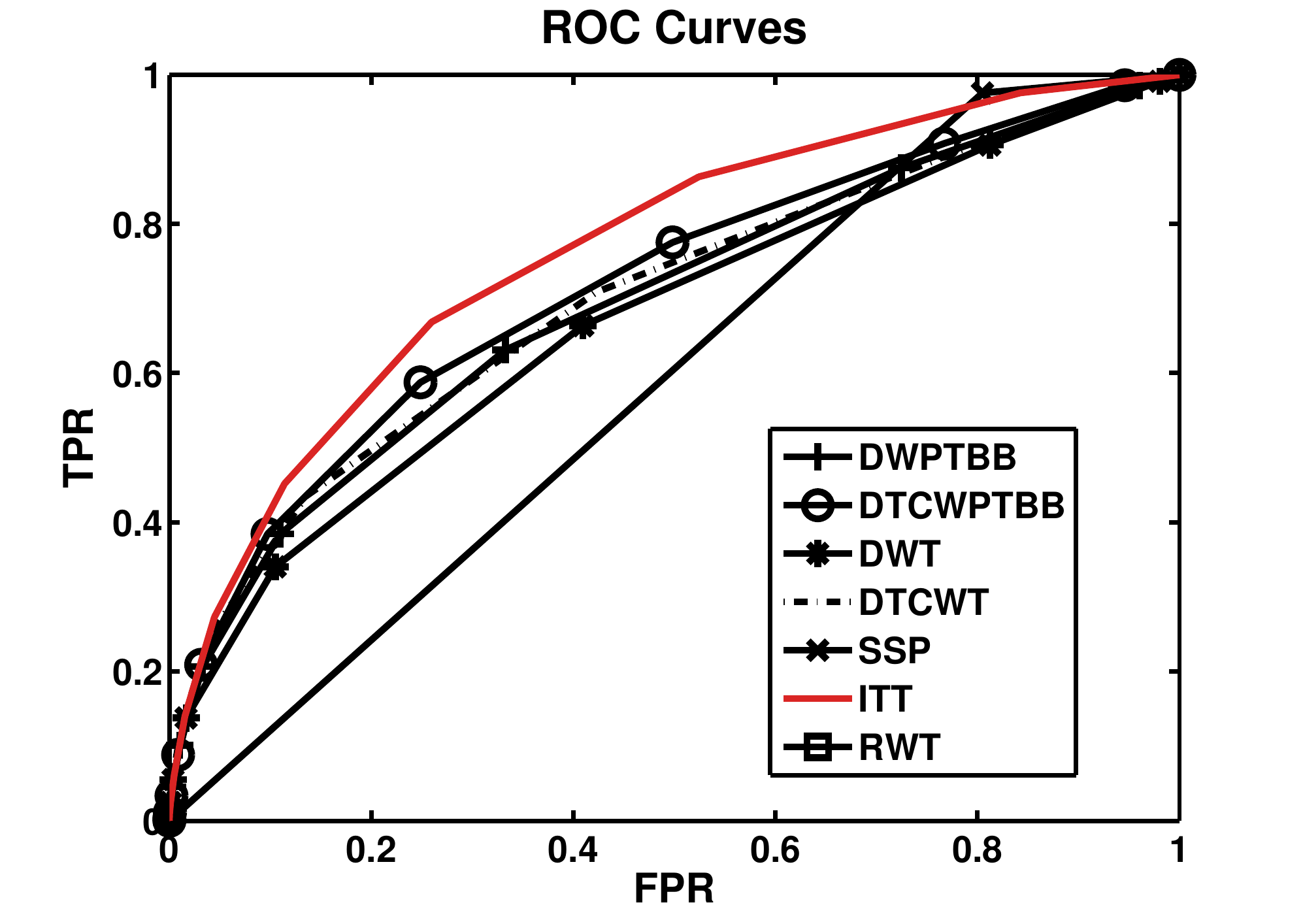}
	\caption{ROC Curve - WSS}%
	\label{fig3:roccuv}
\end{figure}

Above is shown simulation results from Neil Bruce image data-set with eye-tracking locations. Despite of its recently popular database in evaluating saliency maps, the data-set has limitations analysed in the subsection \ref{subsec:database}. Another sets of images should be brought in to enhance diversity of testing samples. Kootstra's database with more image categories and all eye-tracking data ground truth is a perfect candidate. Additional simulation results would confirm and generalize rationale of our proposed information-based saliency methods. Similar to the table \ref{tab1:auctim}, figure \ref{fig1:roccuv}, the table \ref{tab3:auctim} and figure \ref{fig3:roccuv} demonstrate how well the proposed methods with four descriptors and WSS scale selection mechanism perform against other saliency methods like ITT and PSS. 


\begin{table}[!htbp]
	\centering
	\begin{tabular}{cccc}\hline
      	MTH & AUC & NSS & TIME(s) \\ \hline
		ITT & 0.7819 & 0.5144 & 2.3874 \\ 
		PSS & 0.5852 & -0.3532 & 17.0663 \\ 
		DWT & .7150 & 0.4849 & .4414 \\ 
		QWT & .7301 & 0.5070 & 2.9313 \\ 
		DWPTBB & 0.7242 & 0.4631 & 4.9577 \\ 
		QWPTBB & 0.7612 & 0.3922 & 4.7743 \\ \hline	
    	\end{tabular}     
	 \caption{Quantitative Result}%
	 \label{tab3:auctim}
\end{table}

Among four descriptors, the best result in terms of AUC is from QWPTBB descriptor, and the second best is QWT; while both DWT and  DWPTBB have nearly equal AUC values. Comparing with ITT and PSS, QWPTBB's performance in AUC measurement is nearly equal to that of ITT and much larger than PSS. The result strengthens our hypothesis about usefulness of informative clues in saliency map construction. Moreover, it suggest that sub-band wavelet descriptors would be better pixel-based descriptors for scale-saliency computation. In terms of NSS, QWT has slightly out-performed the other descriptors, and its value nearly approaches NSS result of ITT and obviously surpasses PSS's result. 

\begin{figure}[!htbp]
	\centering
	\includegraphics[width=0.75\textwidth]{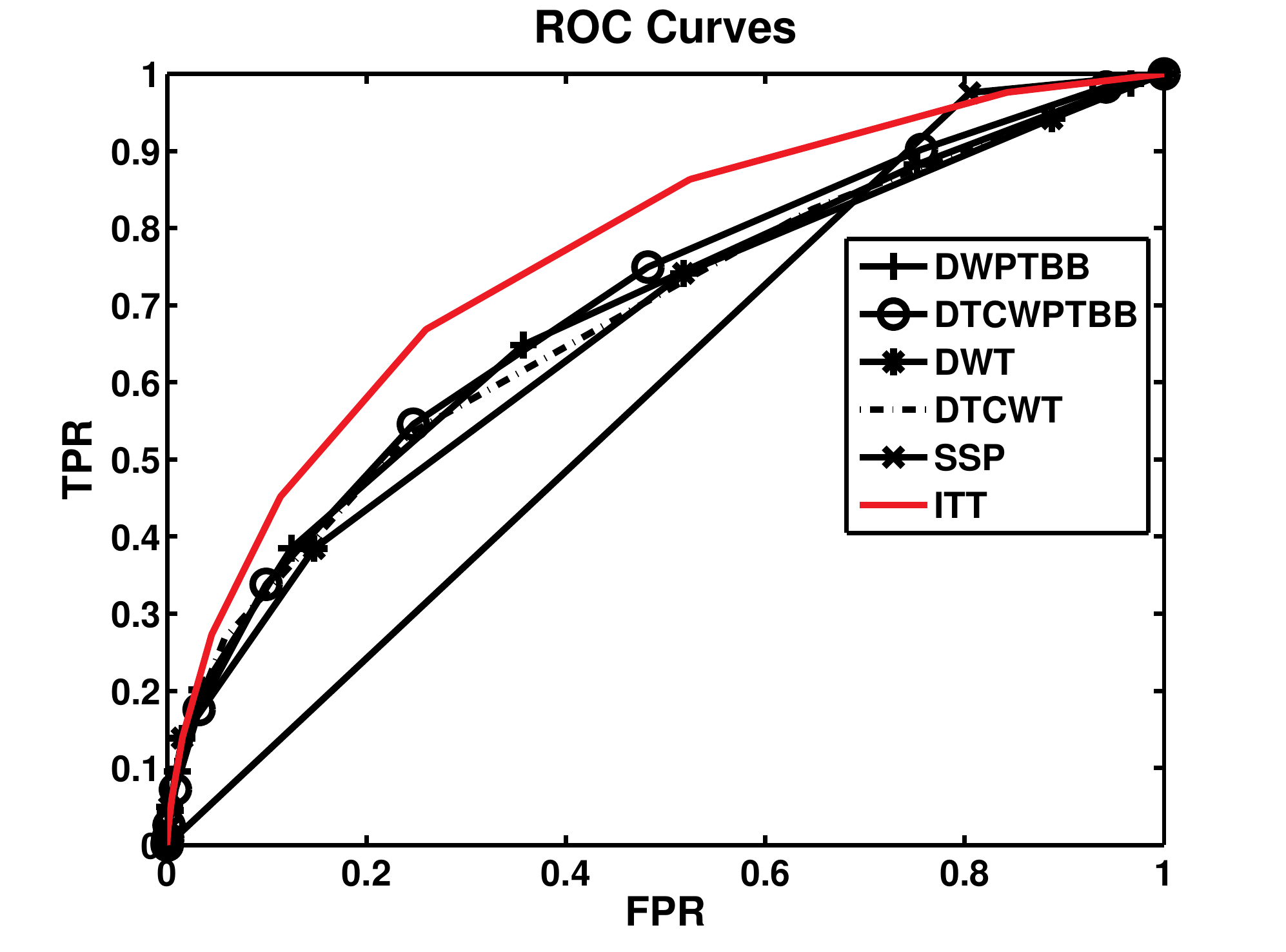}
	\caption{ROC Curve - DIS}%
	\label{fig4:roccuv}
\end{figure}

The graph \ref{fig3:roccuv} and the table \ref{tab3:auctim} shows numeric evaluation of the proposed methods with WSS scale section mechanism on Kootstra's database. Besides WSS scale selection, we have another method called DIS; therefore, we should compare how DIS performs on the same database with suggested sub-band descriptors. Therefore, similar quantitative assessments are also  done for wavelet scale saliency with DIS scale selection and simulation results are shown in figure \ref{fig4:roccuv} and table \ref{tab4:auctim}. 


\begin{table}[!htbp]
	\centering
	\begin{tabular}{cccc}\hline
      	MTH & AUC & NSS & TIME(s) \\ \hline
		ITT & 0.7819 & 0.5144 & 2.3874 \\ 
		PSS & 0.5852 & -0.3532 & 17.0663 \\ 
		DWT & 0.7058 & 0.4847 & 0.4169 \\ 
		DTCWT & 0.7173 & 0.5027 & 2.8972 \\ 
		DWPTBB & 0.7173 & 0.4556 & 4.8156 \\ 
		DTCWPTBB & 0.7381 & 0.3468 & 4.7152 \\ \hline
    	\end{tabular}     
	 \caption{Quantitative Result}%
	 \label{tab4:auctim}
\end{table}

With this specific simulation parameters, the method still performs quite well against other methods like ITT and PSS in both terms of AUC and NSS. However, both AUC and NSS of DIS are slightly worse than those of DIS methods. Noteworthy, ITT analyses three channels color, intensity and orientation simultaneously while we just utilize a intensity channel. Only one channel is chosen since we try to isolate the performance of wavelet scale saliency from other external effects such richness of input features. Regardless of DIS or WSS, the method has very competitive results in numeric terms and it is not due to comprehensiveness of input features.

\subsection{Qualitative Comparison of Saliency Methods}
\label{subsec:result2}
In this section, we show a few examples of visual saliency maps from two mentioned database of Bruce and Kootstra. From each of the databases, only four test images are chosen to be displayed due to limited space though saliency maps are generated for every single image in either of the databases. The samples are intentionally chosen to show variety of contexts and scenes as well as they cover cases of successfully highlighting interested objects and cases of failing to emphasize salient regions. Along with the proposed methods, saliency-maps of ITT and PSS methods are also included so as to give visual comparisons to our proposed saliency methods. Directly below are displayed four samples from Neil Bruce's database.

\begin{figure}[!htbp]
	\centering
	\includegraphics[width=\textwidth,natwidth=1065,natheight=764]{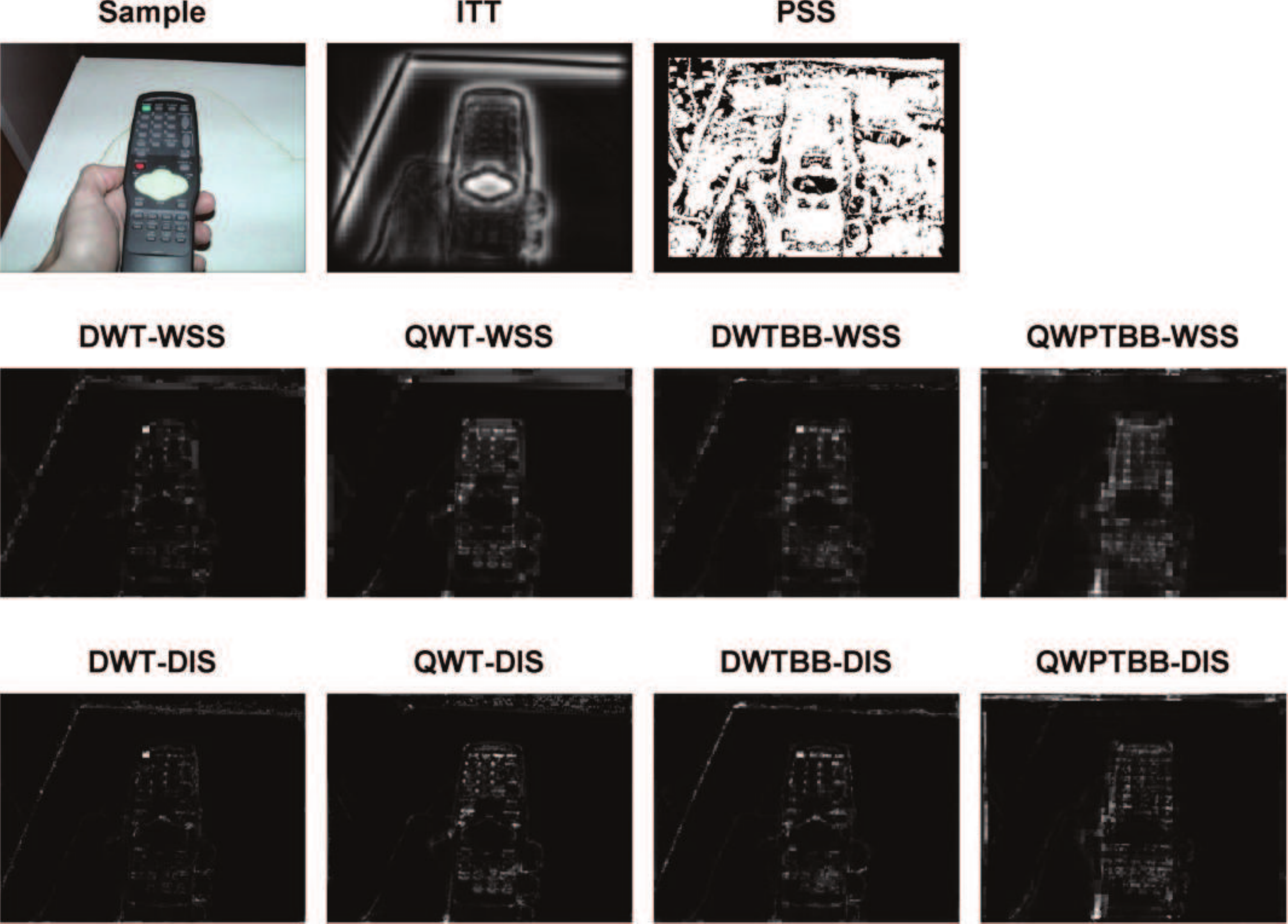}
	\caption{Saliency Map 1}
	\label{fig2:sal1}
\end{figure}

\begin{figure}[!htbp]
	\centering
	\includegraphics[width=\textwidth,natwidth=1065,natheight=764]{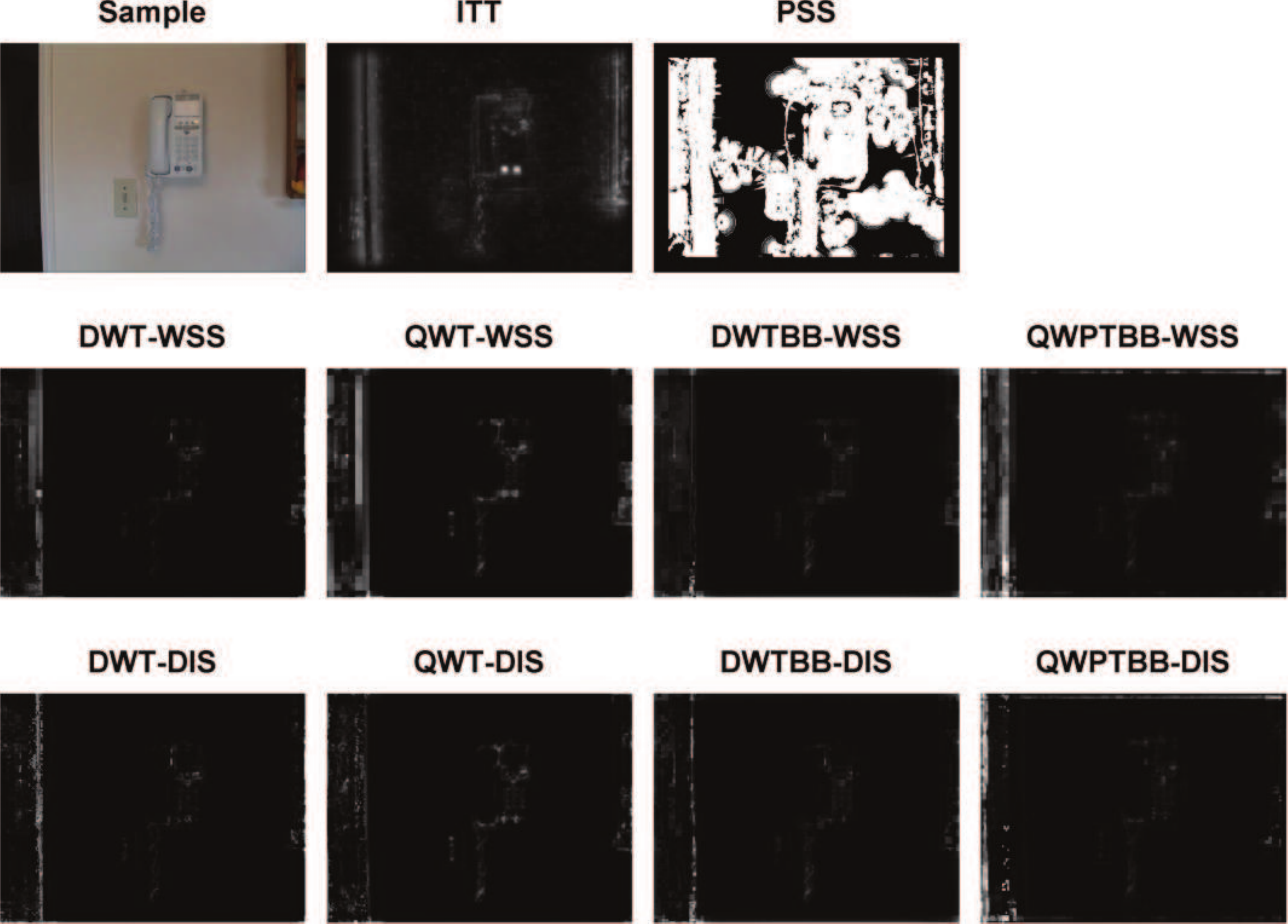}
	\caption{Saliency Map 2}
	\label{fig2:sal2}
\end{figure}

\begin{figure}[!htbp]
	\centering
	\includegraphics[width=\textwidth,natwidth=1065,natheight=764]{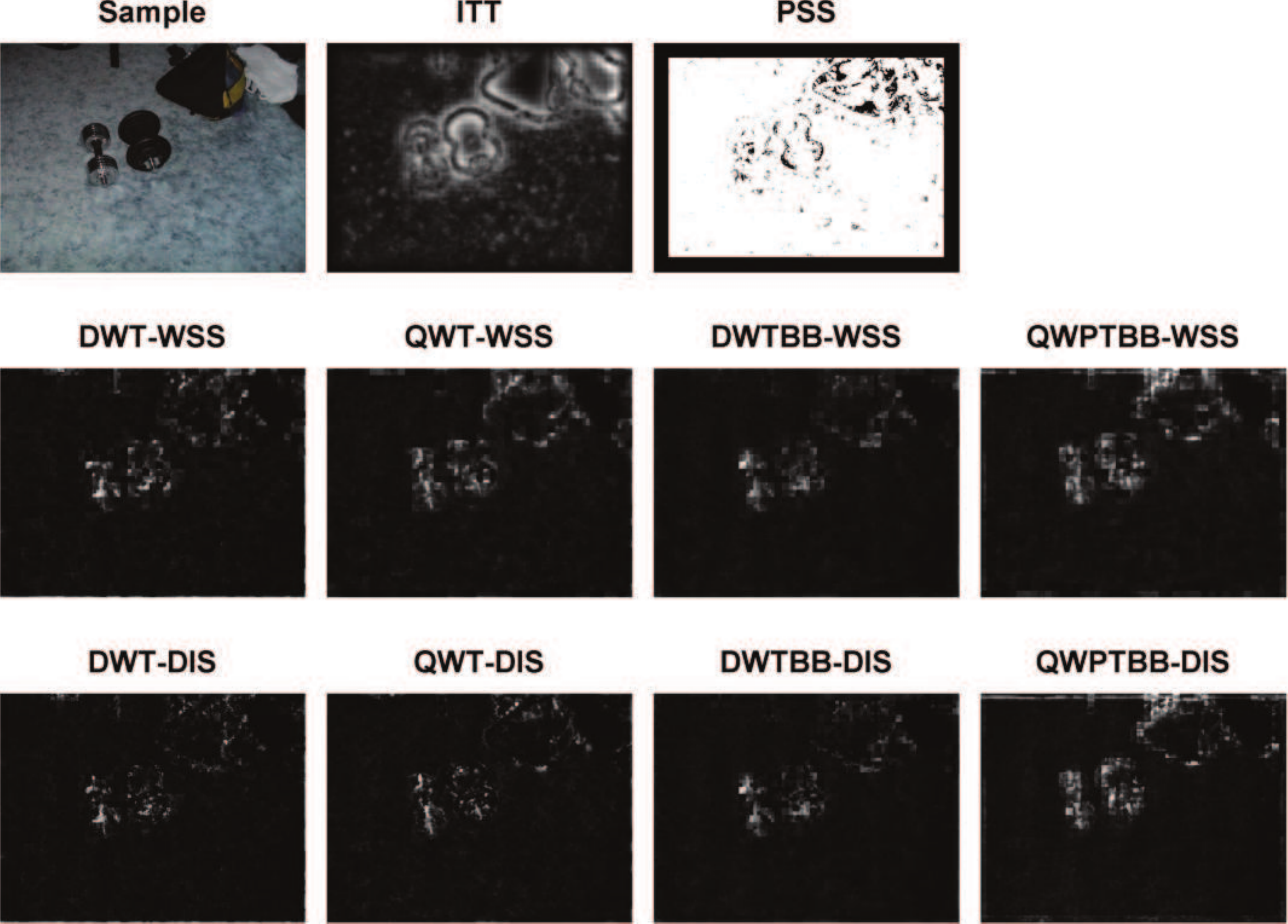}
	\caption{Saliency Map 3}
	\label{fig2:sal3}
\end{figure}

\begin{figure}[!htbp]
	\centering
	\includegraphics[width=\textwidth,natwidth=1065,natheight=764]{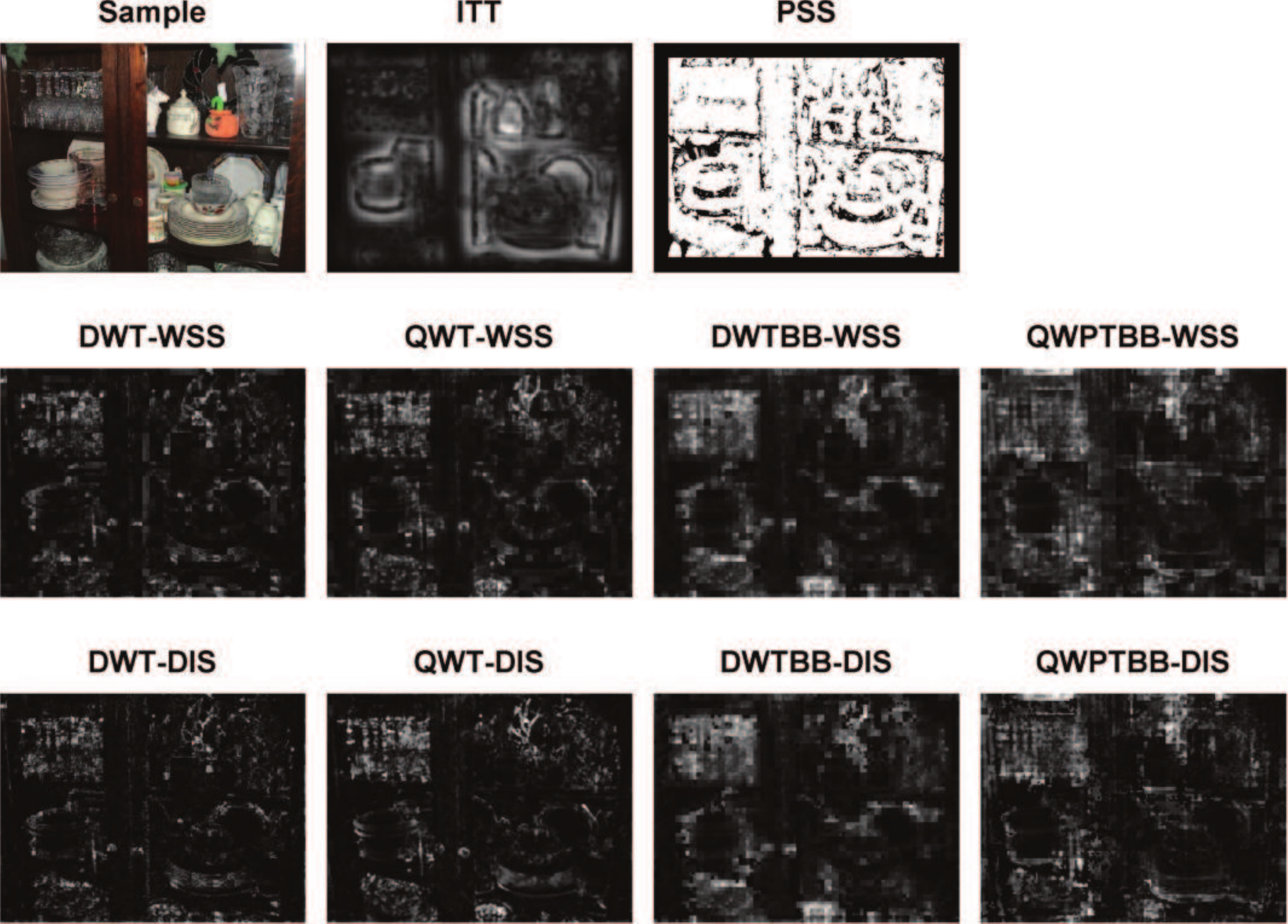}
	\caption{Saliency Map 4}
	\label{fig2:sal4}
\end{figure}

There are four samples , shown in figures \ref{fig2:sal1},\ref{fig2:sal2},\ref{fig2:sal3},\ref{fig2:sal4}, for qualitatively analysing. Generally, PSS identifies a large portion of images as salient regions ( white regions ), it explains why its average AUC and NSS in the table \ref{tab1:auctim} are the lowest, and ITT model gives reasonable saliency maps for three over four samples. Four samples of saliency maps are deliberately chosen to show that different ranking of WSS, DIS derivatives, and their dependence on mother wavelet morphological shapes. Sometimes, their performances are quite similar, the figure \ref{fig2:sal1}; however, QWT-WSS performs better than DWT-WSS in many samples; for example, figure \ref{fig1:roccuv}. DWPTBB-WSS and QWPTBB-WSS also have their own advantages, especially in the case textured background - figure \ref{fig2:sal3}. Finally, sometimes none of proposed methods do give reasonable saliency maps, figure \ref{fig2:sal4}. It usually happens if images are flooded with complex textures.

While Neil Bruce's data-set capture daily scenes in the urban and suburban areas, it lacks of scenes from natural landscapes. Therefore, its images do not represent the whole meaning of "natural images" category. In order to visually confirm effectiveness of our proposed methods, we include four "natural" samples with corresponding saliency maps from the Kootstra's database in the following figures \ref{fig2:sal5}, \ref{fig2:sal6}, \ref{fig2:sal7} and \ref{fig2:sal8}.

\begin{figure}[!htbp]
	\centering
	\includegraphics[width=\textwidth,natwidth=1065,natheight=764]{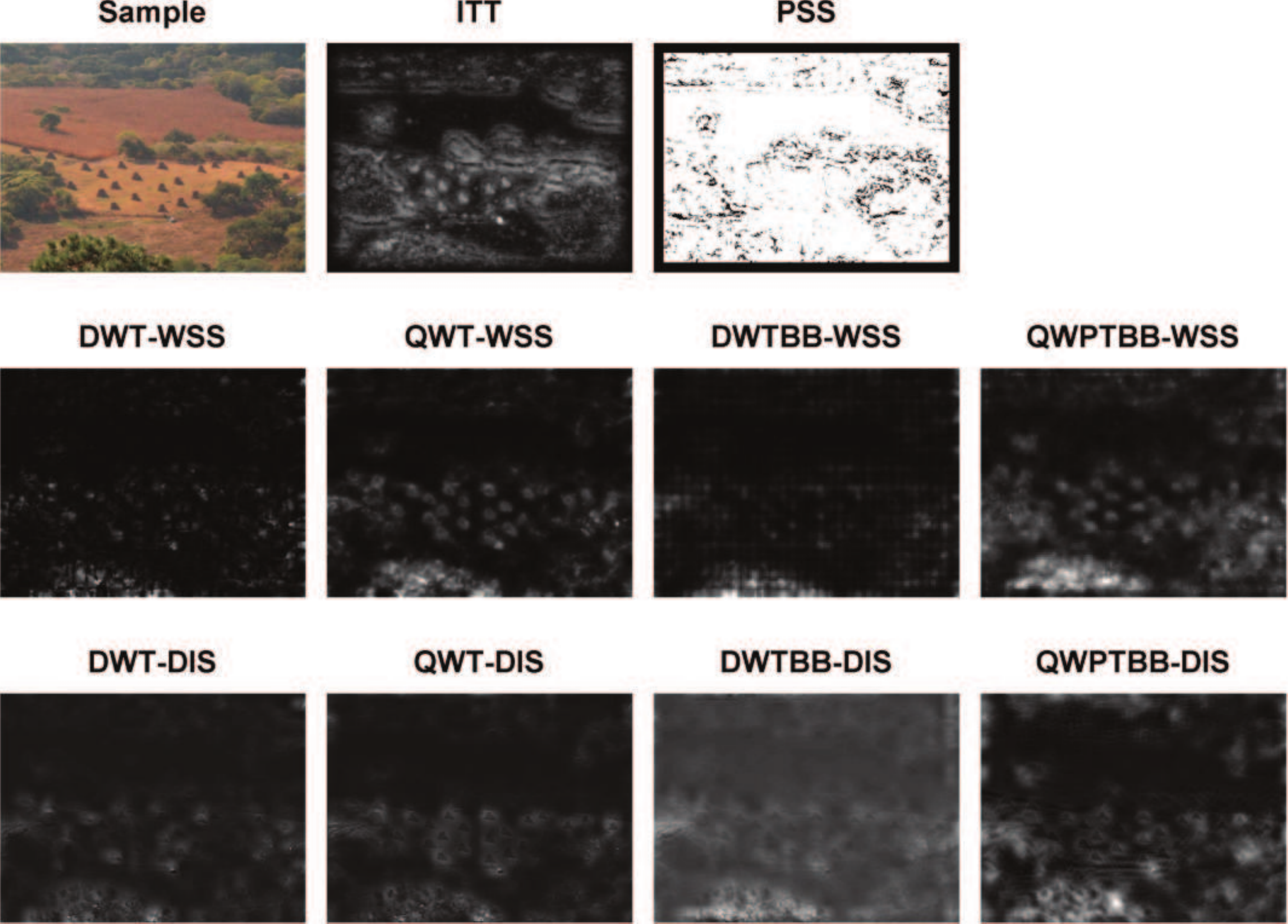}
	\caption{Saliency Map 1}
	\label{fig2:sal5}
\end{figure}
\begin{figure}[!htbp]
	\centering
	\includegraphics[width=\textwidth,natwidth=1065,natheight=764]{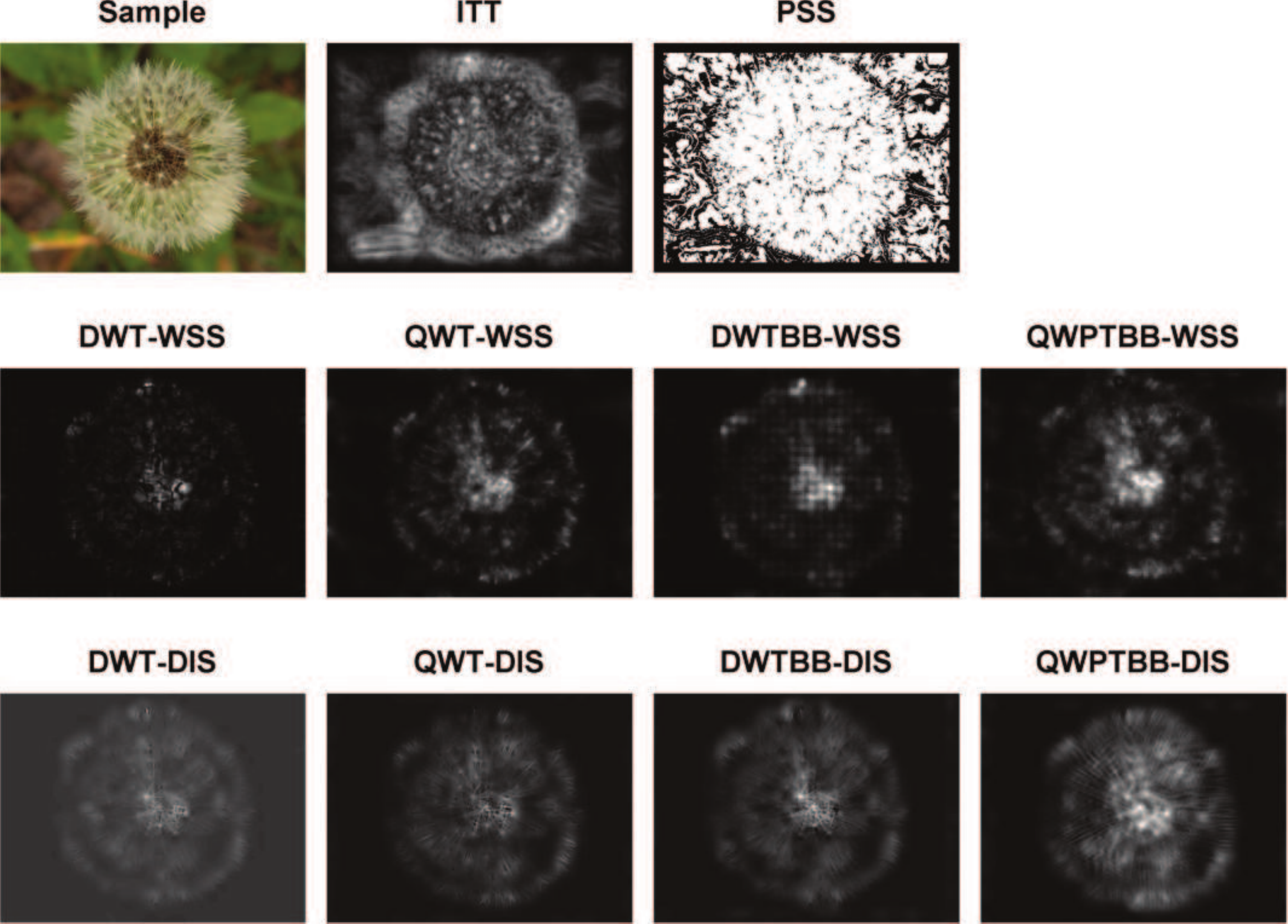}
	\caption{Saliency Map 2}
	\label{fig2:sal6}
\end{figure}
\begin{figure}[!htbp]
	\centering
	\includegraphics[width=\textwidth,natwidth=1065,natheight=764]{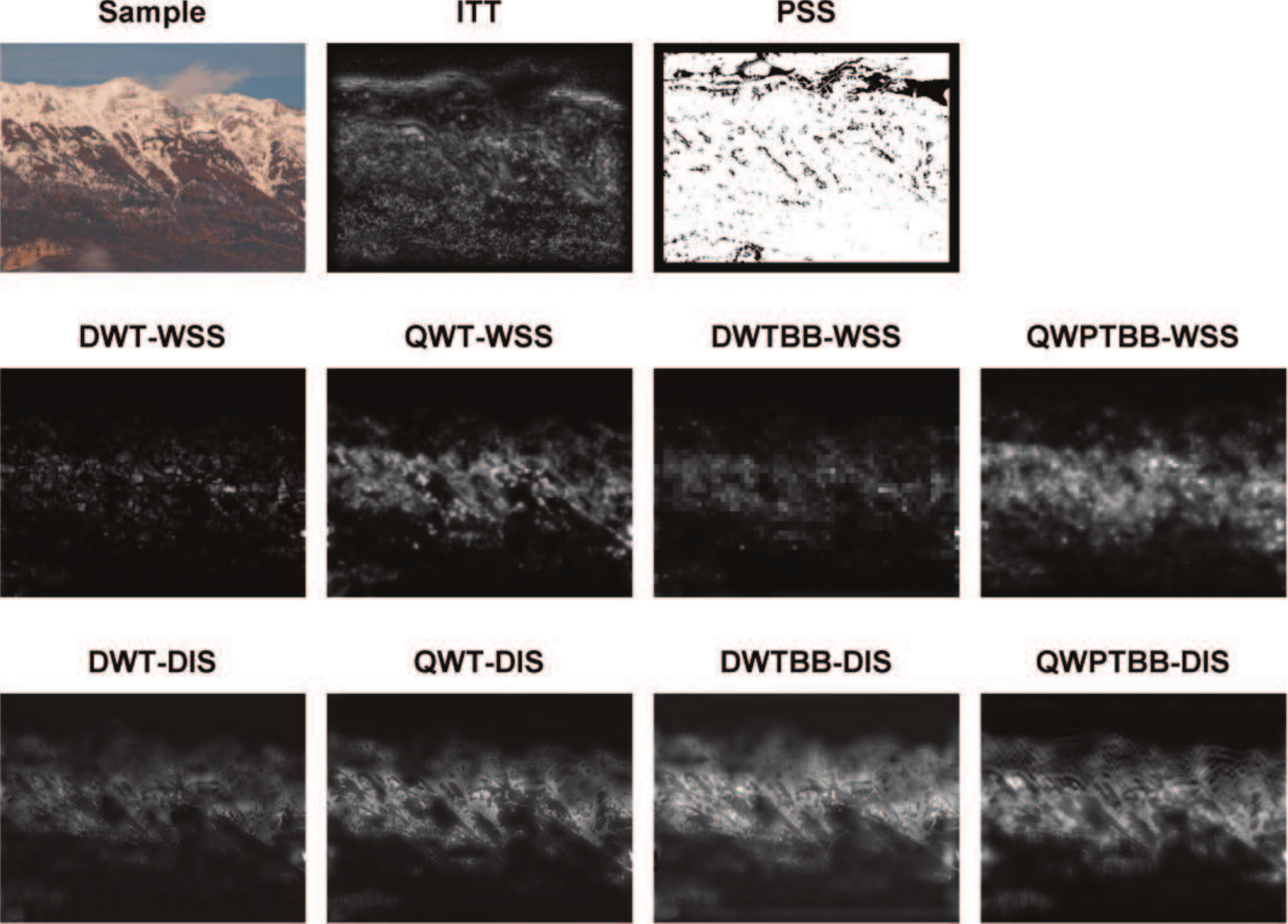}
	\caption{Saliency Map 3}
	\label{fig2:sal7}
\end{figure}
\begin{figure}[!htbp]
	\centering
	\includegraphics[width=\textwidth,natwidth=1065,natheight=764]{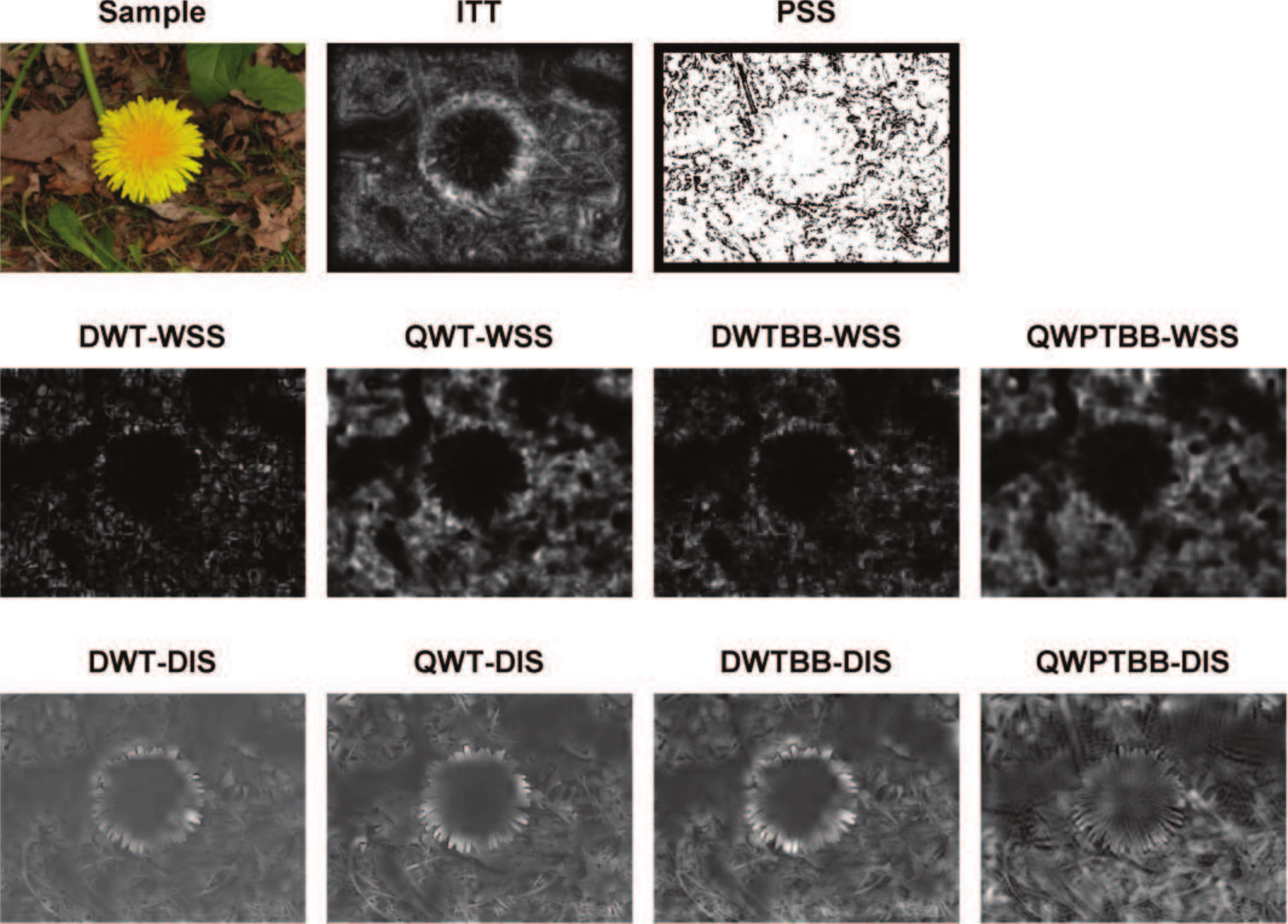}
	\caption{Saliency Map 4}
	\label{fig2:sal8}
\end{figure}

In accordance with one sample image (in color), we display saliency maps produced by ITT , PSS , and eight derivatives of the proposed wavelet scale saliency methods. The figures \ref{fig2:sal5} and \ref{fig2:sal8} represents image of flowers taken in close distance. Therefore, it shows quite a number of symmetric and small details. Meanwhile, figures \ref{fig2:sal6} and \ref{fig2:sal7} contains the whole landscape of mountains and plateaus. Those scenes are usually anti-symmetric and much richer in information than flowery photos. In general observations, ITT method does the best job in selecting the salient features. Among derivatives of the proposed features, QWPTBB descriptors show the most competitive and comprehensive visual result, followed by QWT, DWTBB, and DWT based derivatives in descending order of performances. For comparison between WSS and DIS scale-selection mechanisms, there is slight but significant difference between their saliency maps - WSS saliency maps in the second rows and DIS saliency maps in the third rows of the figures \ref{fig2:sal5},\ref{fig2:sal6},\ref{fig2:sal7}, and \ref{fig2:sal8}. In these examples, DIS maps tend to highlight more features than those of WSS; in other words, WSS maps might have better discriminant power than DIS maps. It would explain why AUC and NSS results in the table \ref{tab3:auctim} are slight better than those in the table \ref{tab4:auctim}. There are small changes in quantitatively visual results when different parameters are used. However, the proposed methods performs very well against other saliency methods like ITT and PSS.

\section{Conclusion}
\label{sec:conclusion}
In this paper, we propose the extension of scale saliency from pixel descriptors to sub-band energy density descriptors generated by four DWT, DWPTBB, QWT, and QWPTBB wavelet transforms with two different scale selection mechanisms WSS and DIS. Comparing to pixel-value descriptors (PSS), the proposed descriptors are much more sparse but biased toward morphological shapes of mother wavelet. Moreover,  the proposed descriptors are more robust to external influencing factors to generation of saliency maps such as shift-variance and other affine transformation. Furthermore, wavelet packet descriptors with best basis algorithms are also considered since several psychological experiments suggest sparseness factor in human vision system. Along with new descriptors, innovative coherent information framework for wavelet scale saliency is proposed and strong relations with Bayesian Surprise Model \cite{Baldi2010} are emphasized. Beside solid theoretical development, the experimental results are as well competitive with state-of-the-art ITT model and surpasses the original scale saliency model PSS quantitatively and qualitatively. In future research, theoretical analysis will be extended to include prior information or top-down information, perceptual grouping and other visual attention operations.

\bibliographystyle{model1-num-names}
\bibliography{JNP_1_2012}

\end{document}